\documentclass{article}

\PassOptionsToPackage{numbers, compress}{natbib}

\usepackage[main, final]{neurips_2026}

\usepackage[utf8]{inputenc} 
\usepackage[T1]{fontenc}    
\usepackage{hyperref}       
\usepackage{url}            
\usepackage{booktabs}       

\usepackage{amsfonts}       
\usepackage{nicefrac}       
\usepackage{microtype}      
\usepackage{xcolor}         
\usepackage[table]{xcolor}
\usepackage{xspace}
\newcommand{\ie}{\emph{i.e.}\xspace}
\newcommand{\eg}{\emph{e.g.}\xspace}
\usepackage{algorithm}
\usepackage{algorithmic}
\usepackage{pifont}
\usepackage{amsmath}
\usepackage{amsfonts}
\usepackage{multirow}
\usepackage{mathabx}

\usepackage{caption}
\usepackage{subcaption}
\usepackage{graphicx}

\usepackage{newfloat}
\usepackage{listings}
\usepackage{soul}       
\sethlcolor{yellow}
\usepackage{booktabs}
\usepackage{multirow}
\usepackage{graphicx}

\usepackage[accsupp]{axessibility}  %

\title{Beyond Token Representations: Explicit Visual Object Grounding for Video Reasoning Segmentation}

\author{%
  \textbf{Jun Huang$^{1}$ \quad
  Junyang Zhang$^{1,2}$ \quad
  Zijie Yue$^{1}$} \quad
  \textbf{Bingkun Wang$^{1}$ \quad
  Yong Luo$^{3}$ \quad
  Miaojing Shi$^{1,4}$}\thanks{Corresponding author.} \\[0.5em]
  $^{1}$College of Electronic and Information Engineering, 
  Tongji University \\
  $^{2}$Shanghai Zhenhua Heavy Industries Co., Ltd. \\
  $^{3}$College of Computer Science, Wuhan University \\
  $^{4}$Shanghai Institute of Intelligent Science and Technology,
Tongji University \\
  \texttt{\{jhuang,zhangjunyang,zijie,wangbingkun,mshi\}@tongji.edu.cn} \\
  \texttt{yong.luo@whu.edu.cn}
}

\begin{document}
\maketitle


\begin{abstract}
Video Reasoning Segmentation (VRS) aims to infer and segment target objects specified by reasoning queries in videos. Existing methods typically first infer the target and generate a few segmentation tokens for it, which are then fed into a mask decoder for mask prediction. However, such segmentation tokens tend to be both semantically monotonous and spatially ambiguous, making them largely insufficient to represent the target to guide mask prediction.
In this paper, we propose an explicit visual object grounding framework for VRS.
Specifically, we first propose a spatial-aware prompt generation and refinement scheme, in which we query a multimodal large language model to generate explicit spatial prompts (\ie, bounding box and point set) for the target and then refine them through multi-step dialogue. Furthermore, we introduce a query diversification and alignment module to generate auxiliary queries that describe the same target from different perspectives. We then enforce cross-query consistency between their predicted spatial prompts to alleviate the training bias caused by query limitations. Finally, we select several reliable frames using a prompt-based keyframe selection strategy to support mask decoding and propagation.
Extensive experiments on eight datasets demonstrate that our method significantly outperforms state-of-the-art methods. The code is available at https://github.com/viscom-tongji/EVOG.

\end{abstract}    
\section{Introduction}

Video Reasoning Segmentation (VRS)~\cite{sdam,videosegr1,vrshq,instructseg} aims to generate segmentation masks for objects specified by reasoning queries in videos.
For example, given queries such as ``the tool used to cut the tomato'' or ``the object that causes the pedestrian to stop before crossing'', the model first implicitly reasons the target object (\ie, a knife or a car) from the video,
then grounds it and generates corresponding pixel-level mask. VRS can be viewed as a more challenging task from Referring Video Object Segmentation (RVOS)~\cite{referdino,losh,weaklyRVOS}. The latter leverages a direct referring paradigm, where the referring expression explicitly specifies the target object via its categorical or visual attributes, \eg, ``the knife cutting the tomato'' or ``the red car in front of the pedestrian''. 

Recent VRS methods~\cite{anyref,lisa,glamm,visa,vrshq} have increasingly adopted Multimodal Large Language Models (MLLMs) to bridge language reasoning and video segmentation.
A seminal work is VISA~\cite{visa}, which employs Chat-UniVi~\cite{chatunivi} to analyze the reasoning query and video content and generates a special segmentation token for the target object. This token is then used as a prompt fed into SAM~\cite{SAM} to predict initial masks for some selected keyframes, which are further propagated to the remaining frames by an object tracker~\cite{xmem}. This paradigm has been widely adopted and extended by subsequent VRS studies~\cite{villa,videolisa,vrshq}. 


Thus far, the key limitation of VRS methods lies in the insufficient representation of target objects, which comes from two aspects: on one hand, the target object is associated with only a few reasoning queries, primarily focusing on its relationship with the environment, thereby leading to a partial observation of the target object. 
Consequently, the model may draw a biased semantic/spatial view on the object in the scene; 
on the other hand, recent methods typically leverage the MLLM to infer the target object into only a few segmentation tokens before feeding them into the mask decoder. As a result, these tokens tend to be both semantically monotonous and spatially ambiguous for the follow-up mask decoder. 
In this paper, we propose an explicit visual object grounding framework for VRS that connects query reasoning and video segmentation through bounding boxes and point sets. 
First, to enhance the spatial indication of the target object beyond the abstract segmentation tokens, we propose a \emph{spatial-aware prompt generation and refinement scheme} that explicitly queries the MLLM to localize the target object in the form of a bounding box and a set of points. Considering that the initial predictions of MLLM may be unreliable, we further refine them through a multi-step dialogue process, in which the MLLM re-evaluates and progressively updates its predictions. This provides more explicit and reliable spatial guidance for the subsequent segmentation decoder. 
Second, to enrich the semantic understanding of the target object beyond the given query context, we introduce a \emph{query diversification and alignment module} that utilizes the MLLM to generate an auxiliary query to inquire the same target from a different perspective; cross-query consistency is then enforced over the generated spatial predictions corresponding to different queries. 
Third, to reduce computational cost, existing VRS methods~\cite{visa,videolisa,vrshq} typically select a small set of keyframes from the video and propagate their masks to remaining frames. Keyframes are determined by their relevance to the reasoning query. Given the generated spatial-aware prompts, we introduce the \emph{prompt-based keyframe selection strategy} by evaluating the geometric correspondence between the box- and point-based prompts on the same object, such that the selected frames are more likely to have the correct target and accurate masks.  

Extensive experiments on VRS dataset ReVOS~\cite{visa} and RVOS datasets including Ref-DAVIS17~\cite{refdavis17}, Ref-YouTube-VOS~\cite{Urvos} and MeViS~\cite{mevis} validate the effectiveness of the proposed framework.

\section{Related Work}

\noindent
\textbf{Referring Video Object Segmentation.}
RVOS~\cite{referdino,RefVOS,CMSA,weaklyRVOS,rethinking} aims to segment target objects in videos based on explicit referring expressions. The success of query-based Transformer architectures in computer vision~\cite{DETR,DeformableDETR} has recently inspired significant advances in RVOS. For example, 
MTTR~\cite{MTTR} adapts the DETR framework~\cite{DETR} to RVOS, while ReferFormer~\cite{ReferFormer} optimizes the Deformable DETR~\cite{DeformableDETR} using fewer instance queries to improve the segmentation efficiency. SgMg~\cite{SgMg} improves ReferFormer~\cite{ReferFormer} by incorporating spectral features to enhance segmentation performance. 
More recently, some methods~\cite{vdit,flowrvs} exploit text-to-video diffusion models for RVOS. For example, VD-IT~\cite{vdit} leverages the rich semantic priors and inherent temporal coherence of latent representations learned from ModelScopeT2V~\cite{modelscope} to promote consistent segmentation across frames.

\noindent
\textbf{Video Reasoning Segmentation.}
Unlike RVOS, where the target object is explicitly specified in the text query, VRS~\cite{decaf,sdam,videosegr1,veasonr1,onlinevrs} requires the model to first infer the target object through reasoning over the input query and video content, and then localize and segment it across frames. For example, VISA~\cite{visa}, the first VRS method, leverages an MLLM to perform cross-modal reasoning and generate a special <SEG> token, which serves as a feature representation for predicting the target object mask. 
HyperSeg~\cite{hyperseg} proposes a temporal adapter in MLLM that performs global prompt aggregation and injects local space-time information to improve temporal understanding for VRS.
VRS-HQ~\cite{vrshq} employs Chat-UniVi~\cite{chatunivi} to generate both frame-level and temporal-level tokens, enabling joint modeling of frame details and temporal dynamics of the target object for segmentation.
However, token-level target representations are inherently abstract and 
tend to be both semantically monotonous and spatially ambiguous,
making it difficult for the segmentation decoder to accurately localize the target object. 

\noindent
\textbf{Multimodal Large Language Models.} 
MLLMs~\cite{Blip2,llava,qwen3,siglip2,vlmhyster} have garnered significant attention due to their impressive performance across a variety of multimodal downstream tasks, such as visual question answering~\cite{llavaonevision} and image captioning~\cite{minigpt}. They learn cross-modal representations by aligning visual and linguistic features through training on a vast corpus of vision-language pairs. Recently, several studies~\cite{villa,visa,videolisa,glus,motion-grounded} have explored the use of MLLMs to solve VRS task. For example, VideoSeg-R1~\cite{videosegr1} introduces reinforcement learning to optimize the reasoning process of an MLLM~\cite{qwen2.5vl}, thereby improving the segmentation performance. 
SDAM~\cite{sdam} leverages an MLLM to analyze object motions in the video and select motion-informative keyframes for mask decoding.

\section{Method}
\begin{figure*}
\centering
\captionsetup{aboveskip=5pt, belowskip=5pt}
\includegraphics[width=\linewidth]
{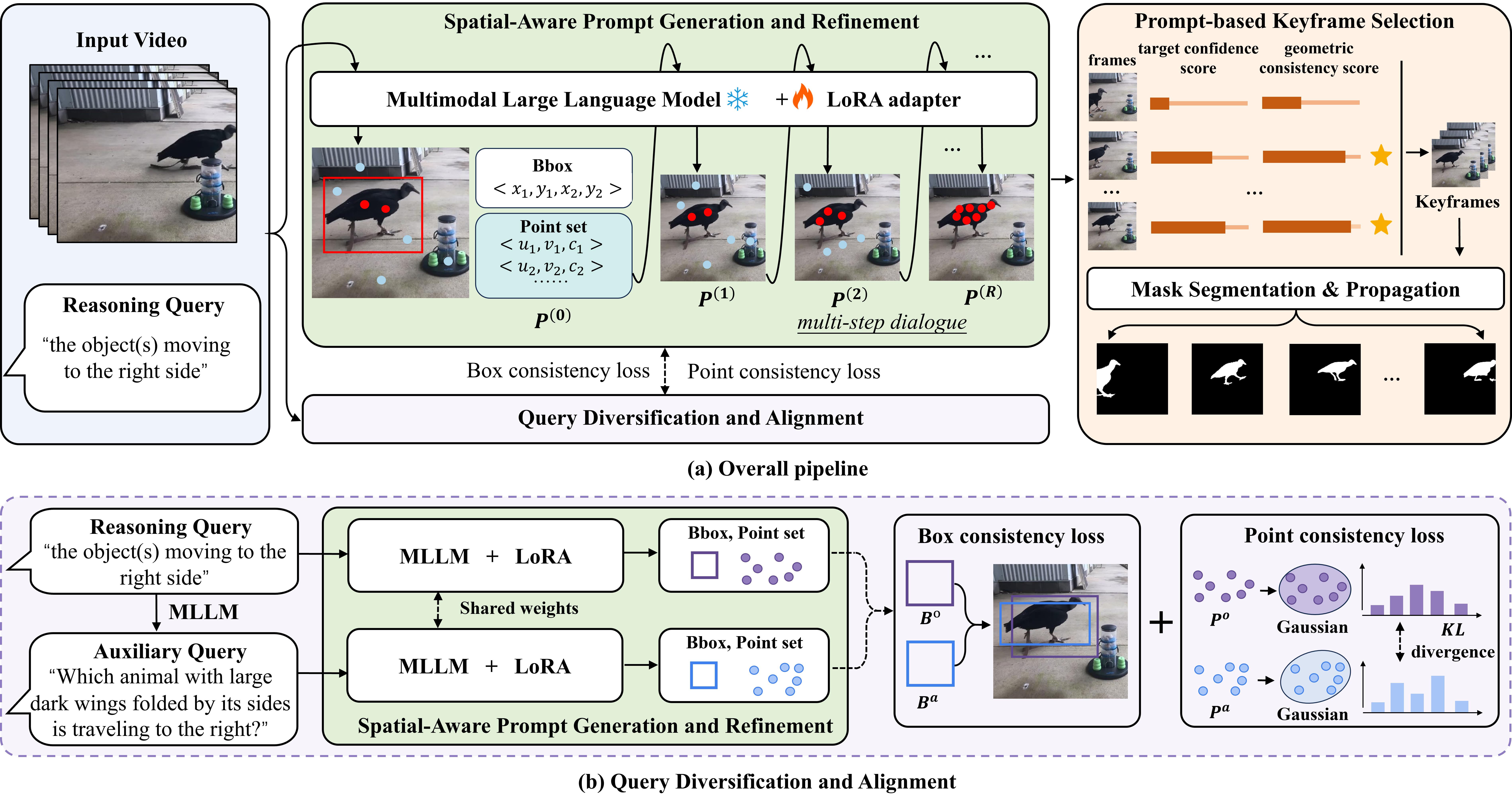}
\caption{Illustration of our proposed framework. (a) overall pipeline; (b) query diversification and alignment module.}
\label{fig:overview}
\vspace{-5mm}
\end{figure*}

Let $\mathcal{V} \in \mathbb{R}^{T \times 3 \times H \times W}$ denote an input video with $T$ frames, and $Q^{o}$ denote the corresponding reasoning query. The objective of VRS is to predict the binary mask sequence $\mathcal{M} \in \mathbb{R}^{T \times H \times W}$ for the target object(s) specified by $Q^{o}$ across frames. 
Figure~\ref{fig:overview} illustrates the framework of our proposed method. We build on the widely used VRS framework~\cite{visa,videolisa,vrshq}, where an MLLM infers the target object from the reasoning query into a few tokens, which are then fed into a mask decoder (\eg, SAM~\cite{SAM} or SAM2~\cite{sam2}) to produce the final segmentation mask. Given the insufficiency of token representations, we offer three new components: spatial-aware prompt generation and refinement scheme, query diversification and alignment module, and prompt-based keyframe selection strategy, to enrich the abstract tokens via explicit spatial-semantic object grounding.  

\subsection{Spatial-aware Prompt Generation and Refinement}
\label{sec:generate-and-refine}
Existing VRS methods typically infer the target object(s) into implicit token-level representations, which provide limited visual information and are difficult to verify or refine before mask decoding. To bridge this gap, we first represent the target object(s) with explicit spatial prompts, \ie, we query the MLLM (\ie, Qwen2.5-VL-7B~\cite{qwen2.5vl}) to predict a bounding box and a sparse set of points for the target object(s). These prompts provide explicit spatial cues and are better aligned with the input format of widely used segmentation models, such as SAM~\cite{SAM} and SAM2~\cite{sam2}, enabling more reliable guidance for mask decoding.


Specifically, given the video $\mathcal{V}$ and the original reasoning query $Q^{o}$, we query the MLLM to generate the spatial prompts using the template:
\par\noindent\begin{minipage}{\linewidth}
\textit{You are given a video and a query that describes the target object(s). Please first localize the target object(s) with a bounding box, and then provide a set of $K$ discriminative points on the target object(s) together with their confidence scores for each frame. Your output must follow the format:} \\
\textit{<BOX> $x_1$ $y_1$ $x_2$ $y_2$ </BOX>; <POINTS> $u_1$ $v_1$ $c_1$; ...; $u_K$ $v_K$ $c_K$ </POINTS>.}
\end{minipage}
\par

Given $\mathcal{V}$, $Q^{o}$, and the above prompt, the MLLM infers the target object and auto-regressively generates the bounding box and point predictions, \ie, a bounding box set $\mathcal{B}=\{B_t\}_{t=1}^{T}$ and a point set $\mathcal{P}=\{P_t\}_{t=1}^{T}$. Here, $B_t=(x_1,y_1,x_2,y_2)$ denotes the bounding box at the $t$-th frame, where $(x_1,y_1)$ and $(x_2,y_2)$ indicate its top-left and bottom-right corners, respectively. The point set at the $t$-th frame is denoted as $P_t=\{p_k\}_{k=1}^{K}$ with $K = 7$, where each point is represented as $p_k=(u_k,v_k,c_k)$. Here, $(u_k,v_k)$ is the 2D coordinate of the $k$-th point, and $c_k\in[0,1]$ denotes its confidence score.

However, the MLLM is typically optimized for semantic understanding rather than spatial localization during their pretraining, \ie, although it may correctly infer which object is the target, it could fail to localize the object. This issue becomes more pronounced in VRS, when the target object is small, partially occluded or temporally ambiguous. To address this issue, we further refine the predictions through a multi-step dialogue process, in which the MLLM is guided to re-evaluate and progressively update its point predictions. Specifically, as shown in Fig.~\ref{fig:overview}, given the video $\mathcal{V}$ and the original query $Q^{o}$, we feed the predicted point set into the MLLM using the following template:
\par\noindent\begin{minipage}{\linewidth}
\textit{You are given a video, a query that describes the target object(s), and a point set with $K$ points predicted in the previous step. Please refine the previous point set by checking whether each point lies on the target object(s) described by the reasoning query. Keep reliable points and replace unreliable points with better suitable points, while maintaining a total of $K$ points for each frame. Your output must follow the format exactly:} \\
\textit{<POINTS> $u_1$ $v_1$ $c_1$; ...; $u_K$ $v_K$ $c_K$ </POINTS>.}
\end{minipage}
\par

Let $\mathcal{P}^{(0)}$ denote the initial point set generated by the MLLM. At the $r$-th refinement step, the MLLM takes $(\mathcal{V}, Q^{o})$ and the point set from the previous step $\mathcal{P}^{(r-1)}$ as input, and produces an updated point set $\mathcal{P}^{(r)}$. 
In practice, we perform $R=4$ refinement steps during training.

To optimize the MLLM's point predictions during iterative steps, we introduce a point calibration loss:
\begin{equation}
\mathcal{L}_{\mathrm{point}}
=
\frac{1}{TR}
\sum_{t=1}^{T}
\sum_{r=1}^{R}
\frac{1}{K}
\sum_{i=1}^{K}
\left[
c_{t,i}^{(r)}(1-m_{t,i}^{(r)}) + (1-c_{t,i}^{(r)})m_{t,i}^{(r)}
\right].
\end{equation}
where $m_{t,i}^{(r)}=1$ if the $i$-th point in the $t$-th frame lies inside the corresponding ground-truth target mask, and $m_{t,i}^{(r)}=0$ otherwise. This loss encourages high confidence for correct points and low confidence for wrong points, thereby reducing false point predictions during iterative refinement. 

Moreover, we follow~\cite{DeformableDETR} to optimize the bounding box predictions using the bounding box loss $\mathcal{L}_{\mathrm{bbox}}$. Notice the bounding box set is empirically chosen to be generated only once.

\subsection{Query Diversification and Alignment}
\label{sec:dqf}
In real-world scenarios, the same target can be described from diverse perspectives, such as appearance, action, interaction, and spatial relation. However, existing VRS datasets typically provide limited reasoning queries primarily describing the target object from a narrow perspective.
As a result, the model may overfit to the specific aspect emphasized by the original query and fail to identify the same target when described from another perspective. To mitigate this issue, we utilize the MLLM to randomly generate an auxiliary query for each target object from an alternative perspective. We then encourage the model to produce consistent predictions for both the original and auxiliary queries.

Specifically, given the video $\mathcal{V}$, the original query $Q^{o}$, and the corresponding ground-truth target mask $\mathcal{M}^{gt}$, we utilize the MLLM to generate an auxiliary query $Q^{a}$ using the following prompt:
\par\noindent\begin{minipage}{\linewidth}
\textit{You are given a video, a reasoning query that describes the target object(s) and the corresponding mask. Generate one additional query that refers to the same target object(s) but from a different perspective, such as:
(1) visual appearance (color, shape, size, texture, distinctive look);
(2) action or interaction (what it is doing, how it interacts with other objects);
(3) spatial relation (its position relative to other objects).}
\end{minipage}
\par

Next, we follow the pipeline described in Sec.~\ref{sec:generate-and-refine} to generate the bounding box set $\mathcal{B}^{a}=\{B_t^{a}\}_{t=1}^{T}$ and point set $\mathcal{P}^{a}=\{P_t^{a}\}_{t=1}^{T}$ for the target object specified by $Q^{a}$. Since both $Q^{a}$ and $Q^{o}$ refer to the same target object, their predictions at each frame, \ie, $B_t^{o}$ and $B_t^{a}$, $P_t^{o}$ and $P_t^{a}$, are expected to be consistent with each other. For bounding box predictions, we minimize a box consistency loss:
\begin{equation}
\mathcal{L}_{\mathrm{b}}
=
\frac{1}{T}
\sum_{t=1}^{T}
\left[
1 - \operatorname{IoU}(B_t^{o}, B_t^{a})
\right].
\end{equation}

While for point set predictions, directly matching individual points between $P_t^{o}$ and $P_t^{a}$ is unrealistic because different queries may activate on different yet valid regions of the same target. 
Instead, we represent each point set as a Gaussian distribution and align the two distributions. It should be noted that, we follow~\cite{probiou} to adopt the single Gaussian as the representation of each point set, with its mean and covariance capturing the overall location. We use Qwen2.5-VL to directly predict the locations of these points and Qwen learns visual-semantic representations from large-scale image-text pairs~\cite{qwen2.5vl}. This suggests that, when prompted to localize the target with a few points, the model is more likely to place them in semantically salient regions of the targets although the targets may involve complex, non-rigid, or articulable objects. The detailed analysis can be found in the appendix.
Specifically, let $P_t^{o}=\{p_k^{o}\}_{k=1}^{K}$ and $P_t^{a}=\{p_k^{a}\}_{k=1}^{K}$.
We compute:
\begin{equation}
\mu_t^{o} = \frac{1}{K}\sum_{k=1}^{K} p_k^{o},
\quad
\Sigma_t^{o}
=
\frac{1}{K}\sum_{k=1}^{K}
(p_k^{o}-\mu_t^{o})(p_k^{o}-\mu_t^{o})^{\top}
+
\epsilon I.
\end{equation}
\begin{equation}
\mu_t^{a} = \frac{1}{K}\sum_{k=1}^{K} p_k^{a},
\quad
\Sigma_t^{a}
=
\frac{1}{K}\sum_{k=1}^{K}
(p_k^{a}-\mu_t^{a})(p_k^{a}-\mu_t^{a})^{\top}
+
\epsilon I.
\end{equation}
where $\epsilon I$ is added for numerical stability, with $I$ denoting the identity matrix. Accordingly, $P_t^{o}$ and $P_t^{a}$ are represented as Gaussian distributions $\mathcal{N}_t^{o}$ and $\mathcal{N}_t^{a}$, parameterized by $(\mu_t^{o}, \Sigma_t^{o})$ and $(\mu_t^{a}, \Sigma_t^{a})$, respectively. We use a point consistency loss to minimize their symmetric KL divergence:
\begin{equation}
\mathcal{L}_{\mathrm{p}}
=
\frac{1}{2T}
\sum_{t=1}^{T}
\left[
D_{\mathrm{KL}}\!\left(\mathcal{N}_t^{o}\middle\|\mathcal{N}_t^{a}\right)
+
D_{\mathrm{KL}}\!\left(\mathcal{N}_t^{a}\middle\|\mathcal{N}_t^{o}\right)
\right].
\label{equation:pts}
\end{equation}

The overall consistency loss is:
\begin{equation}
\mathcal{L}_{\mathrm{c}}
=
\mathcal{L}_{\mathrm{b}}
+
\mathcal{L}_{\mathrm{p}},
\end{equation}
Note that this module is used only during training.



\subsection{Prompt-based Keyframe Selection}
\label{sec:mask-decoding}
To reduce the computational cost, existing VRS methods~\cite{visa,videolisa,vrshq} typically avoid repeatedly applying the mask decoder to every video frame for mask prediction. Instead, they select a small set of keyframes from the input video, predict target masks on these frames and then propagate them through an object tracker (\eg, Cutie~\cite{cutie}) to the remaining frames.
The selected keyframes in these methods are typically based on 
their image-to-text similarities to the reasoning query, yet lack the assessment on the grounded objects in these frames. 


To address this issue, we select keyframes based on two evidence scores: a target confidence score evaluates whether the target object is present in the frame, and a geometric consistency score measures whether the predicted points and bounding box correspond to the same object region. Candidate frames with high average scores are selected as keyframes for mask decoding and propagation.


Specifically, we follow~\cite{vrshq} to leverage SAM2~\cite{sam2} to compute the target confidence score, which indicates the target object's presence in a frame:
\begin{equation}
s_t^{1}
=
{\mathrm{SAM2}}
\left(
\mathcal{V},
B_t^{o},
{P}_t^{o}
\right),
\end{equation}
where $B_t^{o}$ and ${P}_t^{o}$ denote the predicted bounding box and refined point set for the $t$-th frame, respectively. 
A higher $s_t^{1}$ indicates a higher likelihood that the target object is present in the frame.

We then measure the geometric consistency score between the point set and the bounding box:
\begin{equation}
s_t^{2}
=
\frac{1}{|{P}_t^{o}|}
\sum_{(u,v,c)\in {P}_t^{o}}
\mathbb{I}\left[(u,v)\in B_t^{o}\right],
\end{equation}
where $\mathbb{I}[\cdot]$ is the indicator function. This score computes the proportion of points that fall inside the predicted bounding box, with a higher value indicating stronger geometric consistency between the bounding box and point set. For each frame in the video, we compute the above two scores and average them to obtain the final score. The top-$N$ frames with the highest scores are selected as keyframes ($N = 3$).


Next, we follow ~\cite{vrshq,decaf} to utilize SAM2~\cite{sam2} for mask segmentation. For each selected keyframe \(v^{\star}\), we use its corresponding predicted bounding box \(B^{\star}\) and point set \(P^{\star}\) as spatial cues to guide SAM2 to generate the keyframe mask. 
We then leverage Cutie~\cite{cutie} to propagate the keyframe masks to the remaining frames, yielding the final video mask sequence. SAM2 and Cutie are kept frozen during training. Without loss generality, our framework naturally supports multi-instance segmentation when a query refers to multiple target objects. The implementation details are provided in the appendix.

\begin{figure*}
\centering
\captionsetup{aboveskip=5pt, belowskip=5pt}
\includegraphics[width=\linewidth]
{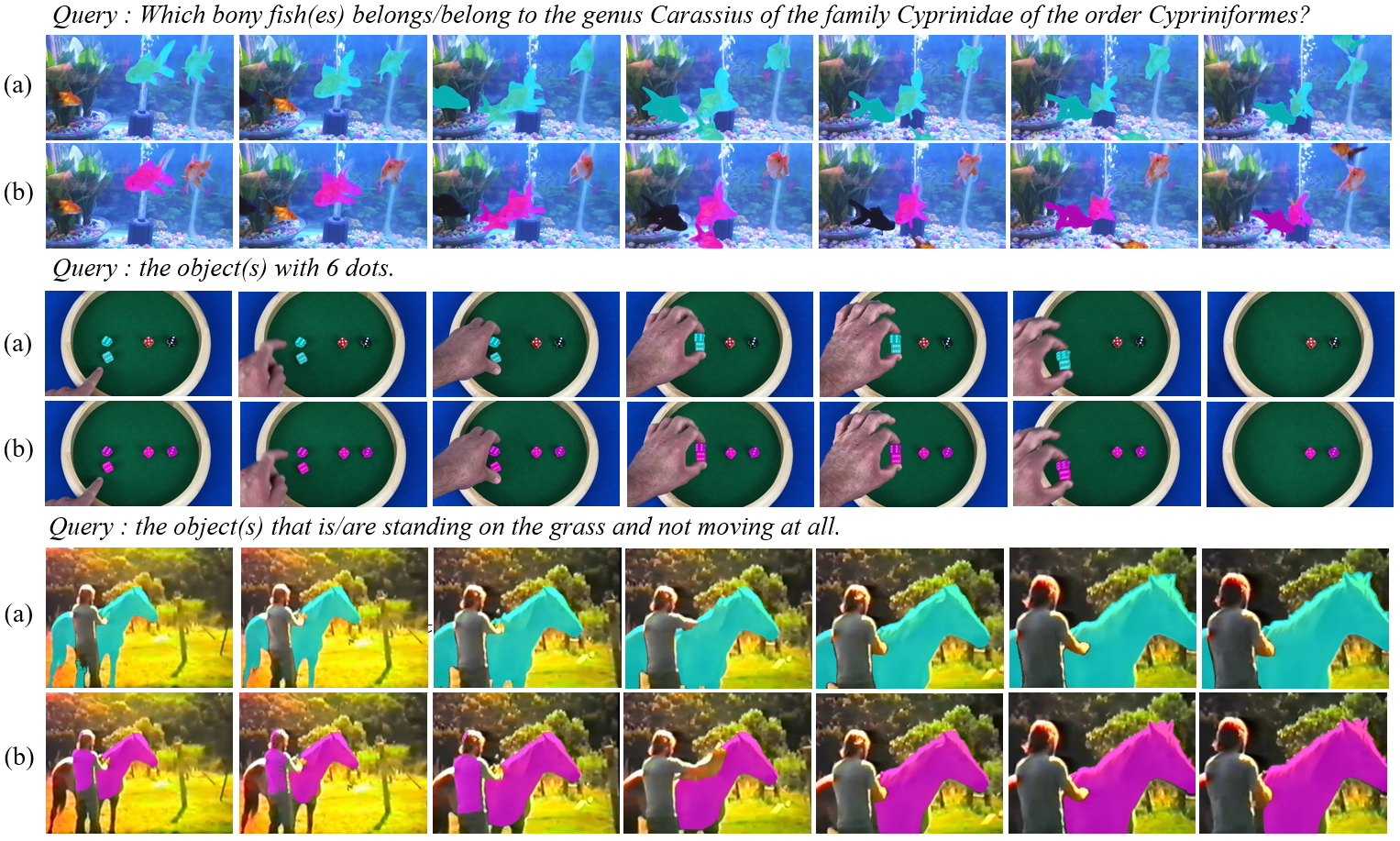}
   \caption{Visualization results on ReVOS. (a) Our method, (b) UniPixel~\cite{unipixel}. }
\label{qualitative_results}
\vspace{-5mm}
\end{figure*}

\subsection{Model Training and Inference}
\label{sec:training-inference}
Our method is optimized in an end-to-end manner using a combination of point calibration loss (Sec.~\ref{sec:generate-and-refine}), bounding box loss (Sec.~\ref{sec:generate-and-refine}), consistency loss (Sec.~\ref{sec:dqf}), and mask loss. The mask loss combines binary cross-entropy (BCE) loss and Dice loss:
\begin{equation}
\mathcal{L}_{\mathrm{mask}}
=
\mathcal{L}_{\mathrm{bce}}({\mathcal{M}},\mathcal{M}^{gt})
+
\mathcal{L}_{\mathrm{dice}}({\mathcal{M}},\mathcal{M}^{gt}),
\label{equation:mask}
\end{equation}
\begin{equation}
\mathcal{L}_{\mathrm{total}}
=
\mathcal{L}_{\mathrm{bbox}}
+
\mathcal{L}_{\mathrm{point}}
+
\mathcal{L}_{\mathrm{mask}}
+
\lambda_{\mathrm{c}}\mathcal{L}_{\mathrm{c}},
\label{equation:total}
\end{equation}
During inference, the model takes only the video $\mathcal{V}$ and the original reasoning query $Q^{o}$ as input. It predicts spatial prompts, refines them through two-step dialogue, selects reliable keyframes, and then decodes and propagates masks. 
\section{Experiments}

\subsection{Experimental Settings}
\label{sec:experimental-settings}
We follow ~\cite{vrshq,videosegr1} to train and evaluate our model on a VRS dataset ReVOS~\cite{visa} and three RVOS datasets, \ie, Ref-DAVIS17~\cite{refdavis17}, Ref-YouTube-VOS~\cite{Urvos} and MeViS~\cite{mevis}. The ReVOS dataset consists of two splits: a referring split for RVOS task with explicit referring expressions and a reasoning split with implicit reasoning queries for VRS task. 
The evaluation metrics are region similarity (\(\mathcal{J}\)), contour accuracy (\(\mathcal{F}\)), and their mean (\(\mathcal{J}\&\mathcal{F}\))~\cite{Urvos}. Moreover, we further evaluate the model on four referring image segmentation datasets, 
\ie, RefCOCO, RefCOCO+, RefCOCOg~\cite{refcoco} and ReasonSeg~\cite{lisa}. The evaluation metrics are
generalized intersection over union (gIoU)~\cite{lisa} and cumulative intersection over union (cIoU)~\cite{lisa}. Notice that the same pipeline in our method applies to VRS, RVOS, and RIS tasks.

\begin{table}[!t]
\centering
\captionsetup{aboveskip=5pt, belowskip=0pt}
\caption{Performance comparison on the ReVOS benchmark. Best results are in \textbf{bold}, and second-best results are \underline{underlined}.}
\label{tab:revos}
\small
\setlength{\tabcolsep}{1.8pt}
\renewcommand{\arraystretch}{1.04}
\begin{tabular*}{\linewidth}{@{\extracolsep{\fill}}lcc ccc ccc ccc@{}}
\toprule
\multirow{2.5}{*}{Method} & \multirow{2.5}{*}{Venue} & \multirow{2.5}{*}{Size}
& \multicolumn{3}{c}{Referring}
& \multicolumn{3}{c}{Reasoning}
& \multicolumn{3}{c}{Overall} \\
\cmidrule(lr){4-6}\cmidrule(lr){7-9}\cmidrule(lr){10-12}
& & & $J$ & $F$ & $J\&F$ & $J$ & $F$ & $J\&F$ & $J$ & $F$ & $J\&F$ \\
\midrule
LISA~\cite{lisa} & CVPR'24 & 7B & 44.3 & 47.1 & 45.7 & 33.8 & 38.4 & 36.1 & 39.1 & 42.7 & 40.9 \\
VISA~\cite{visa} & ECCV'24 & 7B & 49.2 & 52.6 & 50.9 & 40.6 & 45.4 & 43.0 & 44.9 & 49.0 & 46.9 \\
HyperSeg~\cite{hyperseg} & CVPR'25 & 3B & 56.0 & 60.9 & 58.5 & 50.2 & 55.8 & 53.0 & 53.1 & 58.4 & 55.7 \\
VRS-HQ~\cite{vrshq} & CVPR'25 & 7B & 59.8 & 64.5 & 62.1 & 53.5 & 58.7 & 56.1 & 56.6 & 61.6 & 59.1 \\
InstructSeg~\cite{instructseg} & ICCV'25 & 3B & 54.8 & 59.2 & 57.0 & 49.2 & 54.7 & 51.9 & 52.0 & 56.9 & 54.5 \\
ViLLa~\cite{villa} & ICCV'25 & 6B & 57.7 & 61.3 & 59.5 & 52.1 & 56.7 & 54.4 & 54.9 & 59.1 & 57.0 \\
UniPixel~\cite{unipixel} & NeurIPS'25 & 7B & 63.9 & 67.8 & 65.8 & \underline{59.4} & \underline{63.7} & \underline{61.5} & 61.7 & 65.7 & 63.7 \\
Veason-R1~\cite{veasonr1} & arXiv'25 & 7B & 60.7 & 66.5 & 63.6 & 55.8 & 62.2 & 59.0 & 58.2 & 64.4 & 61.3 \\
SDAM~\cite{sdam} & AAAI'26 & 7B & 59.0 & 63.9 & 61.5 & 51.8 & 57.5 & 54.6 & 55.4 & 60.7 & 58.0 \\
VideoSeg-R1~\cite{videosegr1} & AAAI'26 & 7B & 60.5 & 66.3 & 63.4 & 55.5 & 61.9 & 58.7 & 58.2 & 64.0 & 61.1 \\
DecAF~\cite{decaf} & ICLR'26 & 7B & 54.8 & 62.6 & 58.7 & 45.4 & 53.9 & 49.7 & 50.1 & 58.2 & 54.2 \\
\midrule
Ours & -- & 3B & \underline{65.8} & \underline{69.6} & \underline{67.7} & 59.2 & 63.6 & 61.4 & \underline{62.5} & \underline{66.6} & \underline{64.6} \\
Ours & -- & 7B & \textbf{67.5} & \textbf{70.7} & \textbf{69.1} & \textbf{61.5} & \textbf{65.1} & \textbf{63.3} & \textbf{64.5} & \textbf{67.9} & \textbf{66.2} \\
\bottomrule
\end{tabular*}
\vspace{0mm}
\end{table}

\begin{table}[t!]
\centering
\captionsetup{aboveskip=2pt, belowskip=5pt}
\caption{Performance comparison on three RVOS datasets. Best results are in \textbf{bold}, and second-best results are \underline{underlined}.}
\label{tab:rvos}
\small
\setlength{\tabcolsep}{1.8pt}
\renewcommand{\arraystretch}{1.04}
\begin{tabular*}{\linewidth}{@{\extracolsep{\fill}}lcc ccc ccc ccc@{}}
\toprule
\multirow{2.5}{*}{Method} & \multirow{2.5}{*}{Venue} & \multirow{2.5}{*}{Size}
& \multicolumn{3}{c}{MeViS}
& \multicolumn{3}{c}{Ref-YouTube-VOS}
& \multicolumn{3}{c}{Ref-DAVIS17} \\
\cmidrule(lr){4-6}\cmidrule(lr){7-9}\cmidrule(lr){10-12}
& & & $J$ & $F$ & $J\&F$ & $J$ & $F$ & $J\&F$ & $J$ & $F$ & $J\&F$ \\
\midrule
LISA~\cite{lisa} & CVPR'24 & 7B & 35.1 & 39.4 & 37.2 & 53.4 & 54.3 & 53.9 & 62.2 & 67.3 & 64.8 \\
VISA~\cite{visa} & ECCV'24 & 7B & 40.7 & 46.3 & 43.5 & 59.8 & 63.2 & 61.5 & 66.3 & 72.5 & 69.4 \\
VideoLISA~\cite{videolisa} & NeurIPS'24 & 4B & 41.3 & 47.6 & 44.4 & 61.7 & 65.7 & 63.7 & 64.9 & 72.7 & 68.8 \\
HyperSeg~\cite{hyperseg} & CVPR'25 & 3B & - & - & - & 65.7 & 71.3 & 68.5 & 67.9 & 74.5 & 71.2 \\
VRS-HQ~\cite{vrshq} & CVPR'25 & 7B & 47.6 & 53.7 & 50.6 & 68.3 & 72.5 & 70.4 & 72.6 & 79.4 & 76.0 \\
InstructSeg~\cite{instructseg} & ICCV'25 & 3B & -- & -- & -- & 65.4 & 69.5 & 67.5 & 67.3 & 74.9 & 71.1 \\
ViLLa~\cite{villa} & ICCV'25 & 6B & 46.5 & 52.3 & 49.4 & 64.6 & 70.4 & 67.5 & 70.6 & 78.0 & 74.3 \\
UniPixel~\cite{unipixel} & NeurIPS'25 & 7B & 53.2 & 58.3 & 55.8 & 69.5 & 72.4 & 71.0 & 72.7 & 80.1 & 76.4 \\
SDAM~\cite{sdam} & AAAI'26 & 7B & 45.5 & 51.7 & 48.6 & 63.4 & 67.1 & 65.3 & 73.2 & 78.8 & 76.0 \\
VideoSeg-R1~\cite{videosegr1} & AAAI'26 & 7B & 52.7 & 57.8 & 55.3 & \underline{78.2} & \underline{84.4} & \underline{81.3} & \underline{77.4} & \underline{82.2} & \underline{79.8} \\
DecAF~\cite{decaf} & ICLR'26 & 7B & 44.0 & 52.1 & 48.1 & 56.2 & 63.5 & 59.9 & 70.9 & 79.5 & 75.2 \\
\midrule
Ours & -- & 3B & \underline{55.9} & \underline{61.0} & \underline{58.5} & 72.8 & 76.8 & 74.8 & 75.2 & \underline{82.2} & 78.7 \\
Ours & -- & 7B & \textbf{57.7} & \textbf{62.5} & \textbf{60.1} & \textbf{78.4} & \textbf{84.7} & \textbf{81.6} & \textbf{77.5} & \textbf{83.8} & \textbf{80.6} \\
\bottomrule
\end{tabular*}
\vspace{-5mm}
\end{table}

\subsection{Implementation Details}
The default MLLM is Qwen2.5-VL~\cite{qwen2.5vl}, and the mask decoder is SAM2~\cite{sam2}, which align with most existing works~\cite{sdam,videosegr1}. We report results with two MLLM versions, \ie, Qwen2.5-VL-3B and Qwen2.5-VL-7B, to evaluate our method at different MLLM scales.
We apply LoRA~\cite{lora} to the MLLM for parameter-efficient fine-tuning, with the LoRA rank set to 16 and the scaling factor set to 2. The loss weight is set to $\lambda_{\mathrm{c}}=0.2$. 
We utilize the same data loading and preprocessing pipeline as UniPixel~\cite{unipixel}. 
During training, we follow~\cite{ReferFormer} to randomly sample $T=4$ frames per video with a 10-frame interval. 
We optimize the model using the AdamW optimizer with a learning rate of $1e^{-4}$ and a WarmupDecayLR scheduler with 100 warmup iterations. All experiments are conducted using four NVIDIA A40 GPUs.
\subsection{Comparison Results}
As shown in Fig.~\ref{qualitative_results}, our method can accurately reason and segment target objects across frames.

\textbf{VRS.} We first report the model performance on the VRS dataset ReVOS in Tab.~\ref{tab:revos}. We can see our method achieves the best performance across all metrics. In particular, it reaches 69.1\%, 63.3\%, and 66.2\% in $J\&F$ on the referring, reasoning, and overall splits, respectively. Compared with the previous best-performing method, UniPixel~\cite{unipixel},
our method achieves +1.8\% $J\&F$ on the reasoning split and +2.5\% $J\&F$ overall. The gain on the overall split demonstrates that our method not only enhances reasoning segmentation that requires world knowledge and temporal understanding, but also maintains strong generalization ability under the referring setting.



\textbf{RVOS.} Tab.~\ref{tab:rvos} shows that our method achieves the best performance across all RVOS benchmarks. For example, on the MeViS dataset, we achieve 60.1\% in $J\&F$, outperforming the previous best-performing method UniPixel~\cite{unipixel} by +4.3\%. On the Ref-DAVIS17 dataset, we achieve the best results, surpassing VideoSeg-R1~\cite{videosegr1} by 0.8\% in $J\&F$. On the Ref-YouTube-VOS dataset, our method obtains 81.6\% in $J\&F$, outperforming UniPixel by 10.6\% in $J\&F$. 

\textbf{RIS.} We follow ~\cite{visa,unipixel,videosegr1} to seamlessly extend our method to achieve referring image segmentation (RIS) by treating each image as a single-frame video. As shown in Tab.~\ref{tab:image_seg}, 
our method achieves the best results on RefCOCO val/testA, RefCOCO+ testA, RefCOCOg val(U)/test(U) and ReasonSeg gIoU/cIoU splits. For example, compared with the previous best method InstructSeg~\cite{instructseg}, our method improves performance on RefCOCO testA by +1.6\%, RefCOCOg val(U) by +2.2\%, and RefCOCOg test(U) by +2.3\%. Notably, the previous best-performing method InstructSeg~\cite{instructseg} was trained with more data (\ie, plus LLAVA-150k~\cite{llava}) than our method.

\textbf{Computational cost.}
Tab.~\ref{tab:cost} compares our method with existing methods on ReVOS dataset in terms of trainable parameters (TP), GFLOPs, memory usage, and inference speed. Our model has the fewest trainable parameters among the compared methods at 212M, with a computational cost of 1521 GFLOPs and a memory usage of 32.1 GB. It runs at 1.32 FPS, comparable to UniPixel (1.31 FPS), faster than VRS-HQ (1.14 FPS) and VideoSeg-R1 (0.23 FPS), and slower than VISA (1.58 FPS), while achieving better segmentation performance in the reported comparisons as detailed above.
We further examine the cumulative inference cost as components are progressively incorporated into the baseline. Starting from a baseline speed of 2.48 FPS, adding Spatial-aware Prompt Generation and Refinement (SPGR) without point refinement reduces the speed to 2.02 FPS, and enabling point refinement further reduces it to 1.89 FPS. Query Diversification and Alignment (QDA) introduces no inference overhead because it is used only during training. Subsequently adding Prompt-based Keyframe Selection (PKS), SAM2 decoding, and Cutie propagation yields 1.78, 1.46, and 1.32 FPS, respectively. Mask decoding and propagation together account for the largest share of the additional per-frame latency.


\begin{table*}[t!]
\centering
\captionsetup{aboveskip=5pt, belowskip=5pt}
\caption{Performance comparison on four image segmentation datasets. Best results are in \textbf{bold}, and second-best results are \underline{underlined}.}
\label{tab:image_seg}
\footnotesize
\setlength{\tabcolsep}{2.6pt}
\renewcommand{\arraystretch}{1.04}
\resizebox{\textwidth}{!}{%
\begin{tabular}{@{}lcc ccc ccc cc cc@{}}
\toprule
\multirow{2.5}{*}{Method} & \multirow{2.5}{*}{Venue} & \multirow{2.5}{*}{Size}
& \multicolumn{3}{c}{RefCOCO}
& \multicolumn{3}{c}{RefCOCO+}
& \multicolumn{2}{c}{RefCOCOg}
& \multicolumn{2}{c}{ReasonSeg} \\
\cmidrule(lr){4-6}\cmidrule(lr){7-9}\cmidrule(lr){10-11}\cmidrule(lr){12-13}
& & & val & testA & testB & val & testA & testB & val(U) & test(U) & gIoU & cIoU \\
\midrule
LISA~\cite{lisa} & CVPR'24 & 7B & 74.9 & 79.1 & 72.3 & 65.1 & 70.8 & 58.1 & 67.9 & 70.6 & 52.9 & 54.0 \\
PixelLM~\cite{pixellm} & CVPR'24 & 7B & 73.0 & 76.5 & 68.2 & 66.3 & 71.7 & 58.3 & 69.3 & 70.5 & -- & -- \\
GLaMM~\cite{glamm} & CVPR'24 & 7B & 79.5 & 83.2 & 76.9 & 72.6 & 78.7 & 64.6 & 74.2 & 74.9 & -- & -- \\
VISA~\cite{visa} & ECCV'24 & 7B & 72.4 & 75.5 & 68.1 & 59.8 & 64.8 & 53.1 & 65.5 & 66.4 & 52.7 & 57.8 \\
VideoLISA~\cite{videolisa} & NeurIPS'24 & 4B & 73.8 & 76.6 & 68.8 & 63.4 & 68.8 & 56.2 & 68.3 & 68.8 & 61.4 & 67.1 \\
VRS-HQ~\cite{vrshq} & CVPR'25 & 7B & 73.5 & 77.5 & 69.5 & 61.7 & 67.6 & 54.3 & 66.7 & 67.5 & 55.2 & 51.8 \\
HyperSeg~\cite{hyperseg} & CVPR'25 & 3B & 84.8 & 85.7 & 83.4 & 79.0 & 83.5 & \underline{75.2} & 79.4 & 78.9 & 59.2 & 56.7 \\
InstructSeg~\cite{instructseg} & ICCV'25 & 3B & \underline{85.8} & 86.6 & \textbf{84.0} & \textbf{80.1} & \underline{83.8} & \textbf{75.6} & 79.3 & 80.3 & 61.9 & 65.2 \\
UniPixel~\cite{unipixel} & NeurIPS'25 & 7B & 80.8 & 83.0 & 77.4 & 75.3 & 80.1 & 70.0 & 76.4 & 77.1 & 60.5 & 58.7 \\
VideoSeg-R1~\cite{videosegr1} & AAAI'26 & 7B & 78.2 & 82.3 & 75.1 & 71.8 & 76.1 & 64.7 & 73.1 & 74.1 & \underline{69.8} & \underline{68.2} \\
\midrule
Ours & -- & 3B & 84.8 & \underline{86.9} & 79.4 & 76.8 & 82.2 & 71.9 & \underline{80.0} & \underline{81.1} & 65.8 & 64.4 \\
Ours & -- & 7B & \textbf{86.0} & \textbf{88.2} & \underline{83.8} & \underline{79.5} & \textbf{83.9} & 74.1 & \textbf{81.5} & \textbf{82.6} & \textbf{70.1} & \textbf{68.4} \\
\bottomrule
\end{tabular}
}
\vspace{-5mm}
\end{table*}

\subsection{Ablation Studies}
\label{sec:ablation}
We follow ~\cite{unipixel} to conduct ablation studies on the MeViS dataset~\cite{mevis}. 

\noindent\textbf{Spatial-aware Prompt Generation and Refinement.}
\noindent
We evaluate the effectiveness of the proposed spatial-aware prompt generation and refinement (SPGR) module in Tab.~\ref{tab:abl_prompt}. We first propose a variant by removing this module and following VISA~\cite{visa} to use the MLLM to generate token-level target representations to guide mask decoding. This variant leads to a substantial performance drop, \ie, from 60.1\% to 45.5\% in $\mathcal{J}\&\mathcal{F}$. We further evaluate the contribution of each type of spatial prompts, \ie, the bounding box and point set predictions. Removing the point set decreases $\mathcal{J}\&\mathcal{F}$ from 60.1\% to 49.5\%, while removing the bounding box reduces it to 57.4\%. These results indicate that both prompts are beneficial for target grounding and mask decoding.

Furthermore, we query the MLLM to refine the predicted point set through a multi-step dialogue process. Tab.~\ref{tab:abl_refine} reports the model performance under different numbers of refinement steps. When both training and test steps are set to $R=1$, \ie the refinement process is removed, reducing $\mathcal{J}\&\mathcal{F}$ from 60.1\% to 53.7\%. 
We observe that increasing the number of refinement steps during either training or inference consistently improves performance. The performance becomes stable when the training step $R\geq4$ and the inference step $R\geq2$.
In fact, each refinement step brings in about only 5\% additional time, to balance segmentation accuracy and computational efficiency, we use 4 training steps and 2 inference steps as the default setting. 
In addition, we evaluate a variant that also applies multi-step dialogue refinement to bounding box predictions, allowing the MLLM to re-evaluate and update the predicted bounding box. As shown in Tab.~\ref{tab:abl_prompt}, this variant (SPGR w/ box refinement) yields negligible performance gains, suggesting that the initial bounding box already provides sufficient coarse-grained information by capturing the overall extent of the target object.

We assess the box IoU and point-in-mask accuracy of the final MLLM-refined prompts. Box IoU is the IoU between the predicted bounding box and the bounding box derived from the ground truth mask. Point-in-mask accuracy is the percentage of predicted points that fall inside the ground truth mask. As shown in Table~\ref{tab:prompt_quality}, our method achieves 84.63\% Box IoU and 87.32\% point-in-mask accuracy, validating that the refined prompts could reliably localize the target before SAM2.

\begin{table*}[t!]
\centering
\captionsetup{aboveskip=4pt, belowskip=4pt}
%
\begin{minipage}[t]{0.50\textwidth}
\vspace{0pt}
\centering
\captionof{table}{Computational cost and cumulative inference
speed. Rows below Baseline progressively add components.}
\label{tab:cost}
\label{tab:latency_breakdown}
\vspace{2pt}
\footnotesize
\setlength{\tabcolsep}{2.2pt}
\renewcommand{\arraystretch}{1.04}
\begin{tabular*}{\linewidth}{@{\extracolsep{\fill}}lcccc@{}}
\toprule
\textbf{Method}
& \textbf{TP (M)}
& \textbf{GFLOPs}
& \textbf{Memory (GB)}
& \textbf{FPS} \\
\midrule
VISA~\cite{visa}
& 279 & 1340 & 30.5 & 1.58 \\
VRS-HQ~\cite{vrshq}
& 288 & 1765 & 33.9 & 1.14 \\
UniPixel~\cite{unipixel}
& 256 & 1511 & 32.4 & 1.31 \\
VideoSeg-R1~\cite{videosegr1}
& 432 & 6890 & 45.8 & 0.23 \\
\midrule
\multicolumn{4}{@{}l}{Baseline}
& 2.48 \\
\multicolumn{4}{@{}l}{\quad + SPGR w/o point refinement}
& 2.02 \\
\multicolumn{4}{@{}l}{\quad + Point refinement}
& 1.89 \\
\multicolumn{4}{@{}l}{\quad + QDA (training only)}
& 1.89 \\
\multicolumn{4}{@{}l}{\quad + PKS}
& 1.78 \\
\multicolumn{4}{@{}l}{\quad + SAM2 decoding}
& 1.46 \\
\multicolumn{4}{@{}l}{\quad + Cutie propagation (Ours)}
& 1.32 \\
\midrule
Ours
& 212 & 1521 & 32.1 & 1.32 \\
\bottomrule
\end{tabular*}
\end{minipage}%
\hfill
%
\begin{minipage}[t]{0.48\textwidth}
\vspace{0pt}
\centering
\captionsetup{justification=raggedright,singlelinecheck=false}
\footnotesize
\setlength{\tabcolsep}{2.4pt}
\renewcommand{\arraystretch}{1.04}
%
\captionof{table}{Quality of MLLM-generated spatial prompts
before mask decoding.}
\label{tab:prompt_quality}
\vspace{2pt}
\begin{tabular*}{\linewidth}{@{\extracolsep{\fill}}lcc@{}}
\toprule
\textbf{Method}
& \textbf{Box IoU}
& \textbf{Point-in-Mask} \\
\midrule
Ours
& \textbf{84.63} & \textbf{87.32} \\
\bottomrule
\end{tabular*}
\par
\vspace{8pt} 
\captionof{table}{Ablation study on the SPGR module.}
\label{tab:abl_prompt}
\vspace{2pt}
\begin{tabular*}{\linewidth}{@{\extracolsep{\fill}}lccc@{}}
\toprule
\textbf{Method}
& $\mathcal{J}$
& $\mathcal{F}$
& $\mathcal{J}\&\mathcal{F}$ \\
\midrule
Ours w/o SPGR
& 43.0 & 48.1 & 45.5 \\
SPGR w/o point
& 48.7 & 50.3 & 49.5 \\
SPGR w/o box
& 55.6 & 59.2 & 57.4 \\
SPGR w/ box refinement
& \textbf{57.8} & \textbf{62.5} & \textbf{60.1} \\
Ours
& 57.7 & \textbf{62.5} & \textbf{60.1} \\
\bottomrule
\end{tabular*}
\end{minipage}
\par
\vspace{-3mm}
\end{table*}

\begin{table*}[t!]
\centering
\captionsetup{aboveskip=4pt, belowskip=4pt}
\begin{minipage}[t]{0.50\textwidth}
\centering
\captionof{table}{Ablation study on the number of refinement steps in SPGR module. We report $\mathcal{J}\&\mathcal{F}$.}
\label{tab:abl_refine}
\vspace{2pt}
\footnotesize
\setlength{\tabcolsep}{2.4pt}
\renewcommand{\arraystretch}{1.04}
\begin{tabular*}{\linewidth}{@{\extracolsep{\fill}}lccccc@{}}
\toprule
 & \multicolumn{5}{c}{\textbf{Training steps}} \\
\cmidrule(lr){2-6}
\textbf{Test steps} & $R=1$ & $R=2$ & $R=3$ & $R=4$ & $R=5$ \\
\midrule
$R=1$ & 53.7 & 55.4 & 56.1 & 56.3 & 56.4 \\
$R=2$ & 56.4 & 58.2 & 59.6 & 60.1 & 60.1 \\
$R=3$ & 57.4 & 59.6 & 60.0 & 60.1 & \textbf{60.2} \\
$R=4$ & 57.9 & 59.9 & 60.1 & \textbf{60.2} & \textbf{60.2} \\
\bottomrule
\end{tabular*}
\end{minipage}
\hfill
\begin{minipage}[t]{0.48\textwidth}
\centering
\captionof{table}{Ablation study on the QDA module.}
\label{tab:abl_dqf}
\vspace{2pt}
\footnotesize
\setlength{\tabcolsep}{2.2pt}
\renewcommand{\arraystretch}{1.08}
\begin{tabular}{@{}p{0.50\linewidth}ccc@{}}
\toprule
\textbf{Method} & $\mathcal{J}$ & $\mathcal{F}$ & $\mathcal{J}\&\mathcal{F}$ \\
\midrule
Ours w/o QDA & 54.9 & 59.6 & 57.2 \\
Ours w/o $\mathcal{L}_{\mathrm{b}}$ & 56.9 & 61.3 & 59.1 \\
Ours w/o $\mathcal{L}_{\mathrm{p}}$ & 57.2 & 61.6 & 59.4 \\
$\mathcal{L}_{\mathrm{p}}$ $\rightarrow$ $\mathcal{L}_{\mathrm{p\text{-}strict}}$ & 55.2 & 60.0 & 57.6 \\
Ours & \textbf{57.7} & \textbf{62.5} & \textbf{60.1} \\
\bottomrule
\end{tabular}
\end{minipage}
\vspace{-3mm}
\end{table*}

\begin{table*}[t!]
\vspace{2mm}
\centering
\captionsetup{aboveskip=4pt, belowskip=4pt}
\begin{minipage}[t]{0.48\textwidth}
\centering
\captionof{table}{Ablation study on the PKS module.}
\label{tab:abl_keyframe}
\vspace{2pt}
\footnotesize
\setlength{\tabcolsep}{3.0pt}
\renewcommand{\arraystretch}{1.04}
\begin{tabular*}{\linewidth}{@{\extracolsep{\fill}}lccc@{}}
\toprule
\textbf{Method} & $\mathcal{J}$ & $\mathcal{F}$ & $\mathcal{J}\&\mathcal{F}$ \\
\midrule
Ours w/o PKS & 54.6 & 59.4 & 57.0 \\
PKS w/o $s_t^2$ & 56.8 & 61.4 & 59.1 \\
PKS w/o $s_t^1$  & 56.4 & 61.0 & 58.7 \\
SAM3~\cite{sam3}   & 32.3 & 37.5 & 34.9 \\
Ours w/ SAM3  & \textbf{58.0} & \textbf{63.2} & \textbf{60.6} \\
Ours       & 57.7 & 62.5 & 60.1 \\
\bottomrule
\end{tabular*}
\end{minipage}
\hfill
\begin{minipage}[t]{0.48\textwidth}
\centering
\captionof{table}{Ablation study on different MLLMs.}
\label{tab:abl_backbone}
\vspace{2pt}
\footnotesize
\setlength{\tabcolsep}{5.5pt}
\renewcommand{\arraystretch}{1.04}
\begin{tabular}{lccc}
\toprule
\textbf{Backbone} & $\mathcal{J}$ & $\mathcal{F}$ & $\mathcal{J}\&\mathcal{F}$ \\
\midrule
DeepSeek-VL2-3B~\cite{deepseekvl2} & 54.5 & 59.4 & 57.0 \\
DeepSeek-VL2-16B~\cite{deepseekvl2} & \textbf{57.8} & \textbf{62.8} & \textbf{60.3} \\
LLaVA-OneVision-1.5-4B~\cite{llavaonevision1.5} & 55.8 & 61.0 & 58.4 \\
LLaVA-OneVision-1.5-8B~\cite{llavaonevision1.5} & 57.2 & 62.0 & 59.6 \\
Qwen2.5-VL-3B~\cite{qwen2.5vl} & 55.9 & 61.0 & 58.5 \\
Qwen2.5-VL-7B~\cite{qwen2.5vl} & 57.7 & 62.5 & 60.1 \\
\bottomrule
\end{tabular}
\end{minipage}
\vspace{-3mm}
\end{table*}

\noindent\textbf{Query Diversification and Alignment.}
\noindent
We evaluate the effectiveness of the query diversification and alignment (QDA) module in Tab.~\ref{tab:abl_dqf}. Removing this module entirely leads to a clear performance drop, \ie, from 60.1\% to 57.2\% in $\mathcal{J}\&\mathcal{F}$. Moreover, we evaluate two variants that discard one type of consistency constraint. Removing the box-level consistency (Ours w/o $\mathcal{L}_{\mathrm{b}}$) or the point-level consistency (Ours w/o $\mathcal{L}_{\mathrm{p}}$) results in a performance decrease. These results indicate that both box-level and point-level consistencies contribute to QDA, and their combination achieves the best performance. We represent the point sets predicted from the original and auxiliary queries as Gaussian distributions and minimize their symmetric KL divergence. If we instead directly enforce point-wise consistency between the two point sets by minimizing the distance between each point in one set and its nearest point in the other set ($\mathcal{L}_{\mathrm{p}}$ $\rightarrow$ $\mathcal{L}_{\mathrm{p\text{-}strict}}$), the performance drops to 57.6\% in $\mathcal{J}\&\mathcal{F}$. 

In addition, we assess the failure rate of the generated auxiliary queries.
Specifically, we randomly pick 200 samples (each consisting of a video, an original query and an auxiliary query) from ReVOS dataset, then feed each instance into the GPT-5.5 model and ask it to make a binary judgment, \ie it outputs “Yes” only if the two queries refer to the same instance in the video, and “No” otherwise.
As a result, the failure rate of auxiliary queries is only 2\%.

\noindent\textbf{Prompt-based Keyframe Selection.}
\noindent
The prompt-based keyframe selection (PKS) module aims to select reliable frames in terms of a target confidence score and a geometric consistency score. In Tab.~\ref{tab:abl_keyframe}, we denote Ours w/o PKS as a variant that follows ~\cite{visa} to simply select keyframes based on their image-to-text similarities to the reasoning query. This variant decreases $\mathcal{J}\&\mathcal{F}$ from 60.1\% to 57.0\%, indicating that reliable keyframe selection is important for VRS. Moreover, we evaluate two variants that separately use only one of the proposed scores. Using either score alone leads to a performance drop and combining both scores achieves the best.

\noindent\textbf{Choices of SAM series.}
\noindent
We compare our model with a direct deployment of SAM3~\cite{sam3} and a variant of our model in which SAM2 is replaced with SAM3. As shown in Tab.~\ref{tab:abl_keyframe}, SAM3 produces only $34.9\%$ $J\&F$, revealing its limited ability to 
reason the target object from an implicit query. Moreover, replacing SAM2 with SAM3 in our model yields slight performance improvement (from $60.1$\% to $60.6$\% in $\mathcal{J}\&\mathcal{F}$), demonstrating the primary gains of our method arise from developing explicit spatial prompts for visual reasoning segmentation,
rather than from the mask decoder choice.


\noindent\textbf{Choice of MLLMs.}
\noindent
We further evaluate the influence of different MLLM backbones in Tab.~\ref{tab:abl_backbone}. As shown in Tab.~\ref{tab:abl_backbone}, increasing the backbone scale generally leads to better performance. Although DeepSeek-VL2-16B~\cite{deepseekvl2} achieves the best results, we use Qwen2.5-VL-7B as the default backbone because it has comparable results with DeepSeek-VL2-16B using fewer parameters.

\section{Conclusion}
In this paper, we present an explicit visual object grounding framework for video reasoning segmentation. We first propose a spatial-aware prompt generation and refinement module, where the MLLM is queried to infer the target object and explicitly predict its bounding box and point set as spatial prompts. We further design a multi-step dialogue process to progressively update the spatial predictions to guide mask decoding. Moreover, we introduce a query diversification and alignment module that generates auxiliary queries describing the same target object from different perspectives and enforces cross-query consistency between their predicted spatial prompts. Finally, we propose a prompt-based keyframe selection module that selects keyframes based on a target confidence score and a geometric consistency score. This module enables the model to perform mask segmentation on reliable frames, thereby improving the quality of mask propagation. Extensive experiments demonstrate the effectiveness and generalization ability of our method.


\clearpage

\noindent\textbf{Acknowledgments.} This work was supported by the National Natural Science Foundation of China (Grant No.62672351 and No.62401393), the Fundamental Research Funds for the Central Universities, and the Xiaomi Young Scholar Project.

\noindent\textbf{Declaration of competing interests.} The authors declare that they have no known competing financial interests or personal relationships that could have appeared to influence the work reported in this paper.

{
    \small
    \setlength{\parskip}{0pt} 
    \bibliographystyle{ieeenat_fullname}
    \bibliography{main}

@String(CVPR= {IEEE Conf. Comput. Vis. Pattern Recog.})

@String(AAAI = {AAAI})

@String(CVPR  = {CVPR})

@inproceedings{DETR,
  title={End-to-end object detection with transformers},
  author={Carion, Nicolas and Massa, Francisco and Synnaeve, Gabriel and Usunier, Nicolas and Kirillov, Alexander and Zagoruyko, Sergey},
  booktitle={European conference on computer vision},
  pages={213--229},
  year={2020},
  organization={Springer}
}

@Article{DeformableDETR,
  author  = {Zhu, Xizhou and Su, Weijie and Lu, Lewei and Li, Bin and Wang, Xiaogang and Dai, Jifeng},
  journal = {arXiv preprint arXiv:2010.04159},
  title   = {Deformable detr: Deformable transformers for end-to-end object detection},
  year    = {2020},
}

@InProceedings{MTTR,
  author    = {Botach, Adam and Zheltonozhskii, Evgenii and Baskin, Chaim},
  booktitle = {Proceedings of the IEEE/CVF Conference on Computer Vision and Pattern Recognition},
  title     = {End-to-end referring video object segmentation with multimodal transformers},
  year      = {2022},
  pages     = {4985--4995},
}

@InProceedings{ReferFormer,
  author    = {Wu, Jiannan and Jiang, Yi and Sun, Peize and Yuan, Zehuan and Luo, Ping},
  booktitle = {Proceedings of the IEEE/CVF Conference on Computer Vision and Pattern Recognition},
  title     = {Language as queries for referring video object segmentation},
  year      = {2022},
  pages     = {4974--4984},
}

@Article{RefVOS,
  author    = {Bellver, Miriam and Ventura, Carles and Silberer, Carina and Kazakos, Ioannis and Torres, Jordi and Giro-i-Nieto, Xavier},
  journal   = {Multimedia Tools and Applications},
  title     = {A closer look at referring expressions for video object segmentation},
  year      = {2023},
  number    = {3},
  pages     = {4419--4438},
  volume    = {82},
  publisher = {Springer},
}

@Article{rethinking,
  author  = {Liang, Chen and Wu, Yu and Zhou, Tianfei and Wang, Wenguan and Yang, Zongxin and Wei, Yunchao and Yang, Yi},
  journal = {arXiv preprint arXiv:2106.01061},
  title   = {Rethinking cross-modal interaction from a top-down perspective for referring video object segmentation},
  year    = {2021},
}

@inproceedings{Urvos,
  title={Urvos: Unified referring video object segmentation network with a large-scale benchmark},
  author={Seo, Seonguk and Lee, Joon-Young and Han, Bohyung},
  booktitle={European conference on computer vision},
  pages={208--223},
  year={2020},
  organization={Springer}
}

@Article{CMSA,
  author    = {Ye, Linwei and Rochan, Mrigank and Liu, Zhi and Zhang, Xiaoqin and Wang, Yang},
  journal   = {IEEE Transactions on Pattern Analysis and Machine Intelligence},
  title     = {Referring segmentation in images and videos with cross-modal self-attention network},
  year      = {2021},
  number    = {7},
  pages     = {3719--3732},
  volume    = {44},
  publisher = {IEEE},
}

@InProceedings{refcoco,
  author    = {Kazemzadeh, Sahar and Ordonez, Vicente and Matten, Mark and Berg, Tamara},
  booktitle = {Proceedings of the 2014 conference on empirical methods in natural language processing (EMNLP)},
  title     = {Referitgame: Referring to objects in photographs of natural scenes},
  year      = {2014},
  pages     = {787--798},
}

@InProceedings{refdavis17,
  author       = {Khoreva, Anna and Rohrbach, Anna and Schiele, Bernt},
  booktitle    = {Computer Vision--ACCV 2018: 14th Asian Conference on Computer Vision, Perth, Australia, December 2--6, 2018, Revised Selected Papers, Part IV 14},
  title        = {Video object segmentation with language referring expressions},
  year         = {2019},
  organization = {Springer},
  pages        = {123--141},
}

@InProceedings{SgMg,
  author    = {Miao, Bo and Bennamoun, Mohammed and Gao, Yongsheng and Mian, Ajmal},
  booktitle = {Proceedings of the IEEE/CVF International Conference on Computer Vision},
  title     = {Spectrum-guided multi-granularity referring video object segmentation},
  year      = {2023},
  pages     = {920--930},
}

@InProceedings{Blip2,
  author       = {Li, Junnan and Li, Dongxu and Savarese, Silvio and Hoi, Steven},
  booktitle    = {International conference on machine learning},
  title        = {Blip-2: Bootstrapping language-image pre-training with frozen image encoders and large language models},
  year         = {2023},
  organization = {PMLR},
  pages        = {19730--19742},
}

@InProceedings{SAM,
  author    = {Kirillov, Alexander and Mintun, Eric and Ravi, Nikhila and Mao, Hanzi and Rolland, Chloe and Gustafson, Laura and Xiao, Tete and Whitehead, Spencer and Berg, Alexander C and Lo, Wan-Yen and others},
  booktitle = {Proceedings of the IEEE/CVF International Conference on Computer Vision},
  title     = {Segment anything},
  year      = {2023},
  pages     = {4015--4026},
}

@article{qwen3,
  title={Qwen3-vl technical report},
  author={Bai, Shuai and Cai, Yuxuan and Chen, Ruizhe and Chen, Keqin and Chen, Xionghui and Cheng, Zesen and Deng, Lianghao and Ding, Wei and Gao, Chang and Ge, Chunjiang and others},
  journal={arXiv preprint arXiv:2511.21631},
  year={2025}
}

@inproceedings{losh,
  title={Losh: Long-short text joint prediction network for referring video object segmentation},
  author={Yuan, Linfeng and Shi, Miaojing and Yue, Zijie and Chen, Qijun},
  booktitle={Proceedings of the IEEE/CVF Conference on Computer Vision and Pattern Recognition},
  pages={14001--14010},
  year={2024}
}

@article{llavaonevision,
  title={Llava-onevision: Easy visual task transfer},
  author={Li, Bo and Zhang, Yuanhan and Guo, Dong and Zhang, Renrui and Li, Feng and Zhang, Hao and Zhang, Kaichen and Zhang, Peiyuan and Li, Yanwei and Liu, Ziwei and others},
  journal={arXiv preprint arXiv:2408.03326},
  year={2024}
}

@article{siglip2,
  title={Siglip 2: Multilingual vision-language encoders with improved semantic understanding, localization, and dense features},
  author={Tschannen, Michael and Gritsenko, Alexey and Wang, Xiao and Naeem, Muhammad Ferjad and Alabdulmohsin, Ibrahim and Parthasarathy, Nikhil and Evans, Talfan and Beyer, Lucas and Xia, Ye and Mustafa, Basil and others},
  journal={arXiv preprint arXiv:2502.14786},
  year={2025}
}

@inproceedings{referdino,
  title={Referdino: Referring video object segmentation with visual grounding foundations},
  author={Liang, Tianming and Lin, Kun-Yu and Tan, Chaolei and Zhang, Jianguo and Zheng, Wei-Shi and Hu, Jian-Fang},
  booktitle={Proceedings of the IEEE/CVF International Conference on Computer Vision},
  pages={20009--20019},
  year={2025}
}

@article{minigpt,
  title={Minigpt-4: Enhancing vision-language understanding with advanced large language models},
  author={Zhu, Deyao and Chen, Jun and Shen, Xiaoqian and Li, Xiang and Elhoseiny, Mohamed},
  journal={arXiv preprint arXiv:2304.10592},
  year={2023}
}

@article{llava,
  title={Visual instruction tuning},
  author={Liu, Haotian and Li, Chunyuan and Wu, Qingyang and Lee, Yong Jae},
  journal={Advances in neural information processing systems},
  volume={36},
  pages={34892--34916},
  year={2023}
}

@inproceedings{mevis,
  title={Mevis: A large-scale benchmark for video segmentation with motion expressions},
  author={Ding, Henghui and Liu, Chang and He, Shuting and Jiang, Xudong and Loy, Chen Change},
  booktitle={Proceedings of the IEEE/CVF international conference on computer vision},
  pages={2694--2703},
  year={2023}
}

@inproceedings{sam2,
  title={Sam 2: Segment anything in images and videos},
  author={Ravi, Nikhila and Gabeur, Valentin and Hu, Yuan-Ting and Hu, Ronghang and Ryali, Chaitanya and Ma, Tengyu and Khedr, Haitham and R{\"a}dle, Roman and Rolland, Chloe and Gustafson, Laura and others},
  booktitle={International Conference on Learning Representations},
  volume={2025},
  pages={28085--28128},
  year={2025}
}

@inproceedings{anyref,
  title={Multi-modal instruction tuned llms with fine-grained visual perception},
  author={He, Junwen and Wang, Yifan and Wang, Lijun and Lu, Huchuan and He, Jun-Yan and Lan, Jin-Peng and Luo, Bin and Xie, Xuansong},
  booktitle={Proceedings of the ieee/cvf conference on computer vision and pattern recognition},
  pages={13980--13990},
  year={2024}
}

@inproceedings{lisa,
  title={Lisa: Reasoning segmentation via large language model},
  author={Lai, Xin and Tian, Zhuotao and Chen, Yukang and Li, Yanwei and Yuan, Yuhui and Liu, Shu and Jia, Jiaya},
  booktitle={Proceedings of the IEEE/CVF conference on computer vision and pattern recognition},
  pages={9579--9589},
  year={2024}
}

@inproceedings{glamm,
  title={Glamm: Pixel grounding large multimodal model},
  author={Rasheed, Hanoona and Maaz, Muhammad and Shaji, Sahal and Shaker, Abdelrahman and Khan, Salman and Cholakkal, Hisham and Anwer, Rao M and Xing, Eric and Yang, Ming-Hsuan and Khan, Fahad S},
  booktitle={Proceedings of the IEEE/CVF Conference on Computer Vision and Pattern Recognition},
  pages={13009--13018},
  year={2024}
}

@inproceedings{vrshq,
  title={The devil is in temporal token: High quality video reasoning segmentation},
  author={Gong, Sitong and Zhuge, Yunzhi and Zhang, Lu and Yang, Zongxin and Zhang, Pingping and Lu, Huchuan},
  booktitle={Proceedings of the IEEE/CVF Conference on Computer Vision and Pattern Recognition},
  pages={29183--29192},
  year={2025}
}

@inproceedings{villa,
  title={Villa: Video reasoning segmentation with large language model},
  author={Zheng, Rongkun and Qi, Lu and Chen, Xi and Wang, Yi and Wang, Kun and Zhao, Hengshuang},
  booktitle={Proceedings of the IEEE/CVF International Conference on Computer Vision},
  pages={23667--23677},
  year={2025}
}

@inproceedings{decaf,
  title={Decomposed attention fusion in mllms for training-free video reasoning segmentation},
  author={Han, Su Ho and Hyun, Jeongseok and Lee, Pilhyeon and Shim, Minho and Wee, Dongyoon and Kim, Seon Joo},
  booktitle={International Conference on Learning Representations},
  volume={2026},
  pages={28342--28370},
  year={2026}
}

@inproceedings{glus,
  title={Glus: Global-local reasoning unified into a single large language model for video segmentation},
  author={Lin, Lang and Yu, Xueyang and Pang, Ziqi and Wang, Yu-Xiong},
  booktitle={Proceedings of the Computer Vision and Pattern Recognition Conference},
  pages={8658--8667},
  year={2025}
}

@inproceedings{instructseg,
  title={Instructseg: Unifying instructed visual segmentation with multi-modal large language models},
  author={Wei, Cong and Zhong, Yujie and Tan, Haoxian and Zeng, Yingsen and Liu, Yong and Wang, Hongfa and Yang, Yujiu},
  booktitle={Proceedings of the IEEE/CVF International Conference on Computer Vision},
  pages={20193--20203},
  year={2025}
}

@inproceedings{sdam,
  title={Training-Free Spatio-temporal Decoupled Reasoning Video Segmentation with Adaptive Object Memory},
  author={Zhu, Zhengtong and Fan, Jiaqing and Liu, Zhixuan and Li, Fanzhang},
  booktitle={Proceedings of the AAAI Conference on Artificial Intelligence},
  volume={40},
  number={16},
  pages={14022--14030},
  year={2026}
}

@inproceedings{videosegr1,
  title={Videoseg-r1: reasoning video object segmentation via reinforcement learning},
  author={Xu, Zishan and Guo, Yifu and Lu, Yuquan and Yang, Fengyu and Li, Junxin and Cai, Lihua},
  booktitle={Proceedings of the AAAI Conference on Artificial Intelligence},
  volume={40},
  number={14},
  pages={11496--11504},
  year={2026}
}

@inproceedings{visa,
  title={Visa: Reasoning video object segmentation via large language models},
  author={Yan, Cilin and Wang, Haochen and Yan, Shilin and Jiang, Xiaolong and Hu, Yao and Kang, Guoliang and Xie, Weidi and Gavves, Efstratios},
  booktitle={European Conference on Computer Vision},
  pages={98--115},
  year={2024},
  organization={Springer}
}

@article{unipixel,
  title={Unipixel: Unified object referring and segmentation for pixel-level visual reasoning},
  author={Liu, Ye and Ma, Zongyang and Pu, Junfu and Qi, Zhongang and Wu, Yang and Shan, Ying and Chen, Chang},
  journal={Advances in Neural Information Processing Systems},
  volume={38},
  pages={126078--126108},
  year={2026}
}

@article{lora,
  title={Lora: Low-rank adaptation of large language models},
  author={Hu, Edward J and Shen, Yelong and Wallis, Phillip and Allen-Zhu, Zeyuan and Li, Yuanzhi and Wang, Shean and Wang, Lu and Chen, Weizhu},
  journal={arXiv preprint arXiv:2106.09685},
  year={2021}
}

@inproceedings{hyperseg,
  title={Hyperseg: Hybrid segmentation assistant with fine-grained visual perceiver},
  author={Wei, Cong and Zhong, Yujie and Tan, Haoxian and Liu, Yong and Hu, Jie and Li, Dengjie and Zhao, Zheng and Yang, Yujiu},
  booktitle={Proceedings of the Computer Vision and Pattern Recognition Conference},
  pages={8931--8941},
  year={2025}
}

@article{videolisa,
  title={One token to seg them all: Language instructed reasoning segmentation in videos},
  author={Bai, Zechen and He, Tong and Mei, Haiyang and Wang, Pichao and Gao, Ziteng and Chen, Joya and Liu, Lei and Zhang, Zheng and Shou, Mike Z},
  journal={Advances in Neural Information Processing Systems},
  volume={37},
  pages={6833--6859},
  year={2024}
}

@inproceedings{flowrvs,
  title     = {Deforming Videos to Masks: Flow Matching for Referring Video Segmentation},
  author    = {Wang, Zanyi and Jiang, Dengyang and Li, Liuzhuozheng and Dang, Sizhe and Li, Chengzu and Yang, Harry and Dai, Guang and Wang, Mengmeng and Wang, Jingdong},
  booktitle = {International Conference on Learning Representations},
  volume    = {2026},
  pages     = {65586--65601},
  year      = {2026}
}

@inproceedings{chatunivi,
  title={Chat-univi: Unified visual representation empowers large language models with image and video understanding},
  author={Jin, Peng and Takanobu, Ryuichi and Zhang, Wancai and Cao, Xiaochun and Yuan, Li},
  booktitle={Proceedings of the IEEE/CVF Conference on Computer Vision and Pattern Recognition},
  pages={13700--13710},
  year={2024}
}

@article{llavaonevision1.5,
  title={Llava-onevision-1.5: Fully open framework for democratized multimodal training},
  author={An, Xiang and Xie, Yin and Yang, Kaicheng and Zhang, Wenkang and Zhao, Xiuwei and Cheng, Zheng and Wang, Yirui and Xu, Songcen and Chen, Changrui and Zhu, Didi and others},
  journal={arXiv preprint arXiv:2509.23661},
  year={2025}
}

@article{deepseekvl2,
  title={Deepseek-vl2: Mixture-of-experts vision-language models for advanced multimodal understanding},
  author={Wu, Zhiyu and Chen, Xiaokang and Pan, Zizheng and Liu, Xingchao and Liu, Wen and Dai, Damai and Gao, Huazuo and Ma, Yiyang and Wu, Chengyue and Wang, Bingxuan and others},
  journal={arXiv preprint arXiv:2412.10302},
  year={2024}
}

@inproceedings{vdit,
  title={Exploring pre-trained text-to-video diffusion models for referring video object segmentation},
  author={Zhu, Zixin and Feng, Xuelu and Chen, Dongdong and Yuan, Junsong and Qiao, Chunming and Hua, Gang},
  booktitle={European Conference on Computer Vision},
  pages={452--469},
  year={2024},
  organization={Springer}
}

@article{qwen2.5vl,
  title   = {Qwen2.5-VL Technical Report},
  author  = {Bai, Shuai and Chen, Keqin and Liu, Xuejing and Wang, Jialin and Ge, Wenbin and Song, Sibo and Dang, Kai and Wang, Peng and Wang, Shijie and Tang, Jun and Zhong, Humen and Zhu, Yuanzhi and Yang, Mingkun and Li, Zhaohai and Wan, Jianqiang and Wang, Pengfei and Ding, Wei and Fu, Zheren and Xu, Yiheng and Ye, Jiabo and Zhang, Xi and Xie, Tianbao and Cheng, Zesen and Zhang, Hang and Yang, Zhibo and Xu, Haiyang and Lin, Junyang},
  journal = {arXiv preprint arXiv:2502.13923},
  year    = {2025},
  doi     = {10.48550/arXiv.2502.13923},
  url     = {https://arxiv.org/abs/2502.13923}
}

@inproceedings{xmem,
  title={Xmem: Long-term video object segmentation with an atkinson-shiffrin memory model},
  author={Cheng, Ho Kei and Schwing, Alexander G},
  booktitle={European conference on computer vision},
  pages={640--658},
  year={2022},
  organization={Springer}
}

@article{modelscope,
  title={Modelscope text-to-video technical report},
  author={Wang, Jiuniu and Yuan, Hangjie and Chen, Dayou and Zhang, Yingya and Wang, Xiang and Zhang, Shiwei},
  journal={arXiv preprint arXiv:2308.06571},
  year={2023}
}

@article{veasonr1,
  title={Reinforcing video reasoning segmentation to think before it segments},
  author={Gong, Sitong and Zhang, Lu and Zhuge, Yunzhi and Jia, Xu and Zhang, Pingping and Lu, Huchuan},
  journal={arXiv preprint arXiv:2508.11538},
  year={2025}
}

@inproceedings{onlinevrs,
  title={Online reasoning video segmentation with just-in-time digital twins},
  author={Shen, Yiqing and Liu, Bohan and Li, Chenjia and Seenivasan, Lalithkumar and Unberath, Mathias},
  booktitle={Proceedings of the IEEE/CVF International Conference on Computer Vision},
  pages={24698--24706},
  year={2025}
}

@inproceedings{motion-grounded,
  title={Motion-grounded video reasoning: Understanding and perceiving motion at pixel level},
  author={Deng, Andong and Chen, Tongjia and Yu, Shoubin and Yang, Taojiannan and Spencer, Lincoln and Tian, Yapeng and Mian, Ajmal Saeed and Bansal, Mohit and Chen, Chen},
  booktitle={Proceedings of the Computer Vision and Pattern Recognition Conference},
  pages={8625--8636},
  year={2025}
}

@inproceedings{sam3,
  title={Sam 3: Segment anything with concepts},
  author={Carion, Nicolas and Gustafson, Laura and Hu, Yuan-Ting and Debnath, Shoubhik and Hu, Ronghang and Suris Coll-Vinent, Didac and Ryali, Chaitanya and Alwala, Kalyan Vasudev and Khedr, Haitham and Huang, Andrew and others},
  booktitle={International conference on learning representations},
  volume={2026},
  pages={138846--138923},
  year={2026}
}

@article{kld,
  title={Learning high-precision bounding box for rotated object detection via kullback-leibler divergence},
  author={Yang, Xue and Yang, Xiaojiang and Yang, Jirui and Ming, Qi and Wang, Wentao and Tian, Qi and Yan, Junchi},
  journal={Advances in Neural Information Processing Systems},
  volume={34},
  pages={18381--18394},
  year={2021}
}

@article{probiou,
  title={Probabilistic intersection-over-union for training and evaluation of oriented object detectors},
  author={Murrugarra-Llerena, Jeffri and Kirsten, Lucas N and Zeni, Luis Felipe and Jung, Claudio R},
  journal={IEEE Transactions on Image Processing},
  volume={33},
  pages={671--681},
  year={2024},
  publisher={IEEE}
}

@inproceedings{cutie,
  title={Putting the object back into video object segmentation},
  author={Cheng, Ho Kei and Oh, Seoung Wug and Price, Brian and Lee, Joon-Young and Schwing, Alexander},
  booktitle={2024 IEEE/CVF Conference on Computer Vision and Pattern Recognition (CVPR)},
  pages={3151--3161},
  year={2024},
  organization={IEEE}
}

@article{weaklyRVOS,
  title={Weakly-Supervised Referring Video Object Segmentation through Text Supervision},
  author={Shi, Miaojing and Huang, Jun and Yue, Zijie and Wang, Hanli},
  journal={arXiv preprint arXiv:2604.17797},
  year={2026}
}

@article{vlmhyster,
  title={Bootstrapping vision-language model for hysteroscopic surgical scene segmentation},
  author={Huang, Jun and Chen, Meiyi and Yue, Zijie and Xiao, Yuhang and Li, Fang and Wang, Hanli and Tong, Xiaowen and Guo, Yi and Shi, Miaojing},
  journal={Biomedical Signal Processing and Control},
  volume={127},
  pages={111214},
  year={2026},
  publisher={Elsevier}
}

@inproceedings{pixellm,
  title={Pixellm: Pixel reasoning with large multimodal model},
  author={Ren, Zhongwei and Huang, Zhicheng and Wei, Yunchao and Zhao, Yao and Fu, Dongmei and Feng, Jiashi and Jin, Xiaojie},
  booktitle={2024 IEEE/CVF Conference on Computer Vision and Pattern Recognition (CVPR)},
  pages={26364--26373},
  year={2024},
  organization={IEEE}
}
}

\appendix
\clearpage
\setcounter{section}{0}
\setcounter{figure}{0}
\setcounter{table}{0}
\renewcommand{\thesection}{S\arabic{section}}
\renewcommand{\thefigure}{S\arabic{figure}}
\renewcommand{\thetable}{S\arabic{table}}

\section*{Overview of Appendix}
This supplementary material provides 1) additional ablation studies, 2) more visualization results, 3) failure cases.

\section{Additional Ablation Studies}
\subsection{Spatial-aware Prompt Generation and Refinement.}

\begin{figure*}[!t]
\centering
\captionsetup{aboveskip=10pt, belowskip=5pt}
\includegraphics[width=\linewidth]
{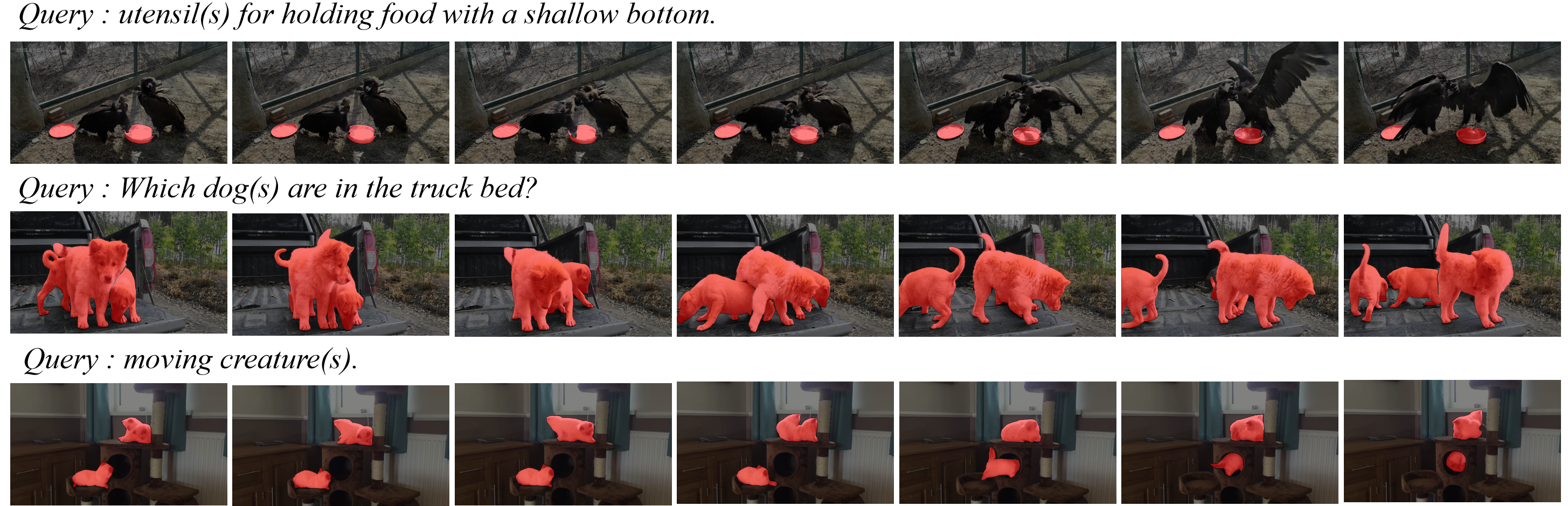}
   \caption{Visualization results of some multi-instance cases. }
\label{multiinstance}
\vspace{-5mm}
\end{figure*}

\begin{table}[t]
\centering
\caption{Detailed analysis on the ReVOS test set.}
\label{tab:revos_detailed}
\begin{tabular}{lcc}
\toprule
Subset &  Samples &  $\mathcal{J}\&\mathcal{F}$ \\
\midrule
Single-instance & 4,719 & 66.5 \\
Multi-instance  & 880   & 64.6 \\
\midrule
Overall        & 5,599 & 66.2 \\
\bottomrule
\end{tabular}
\end{table}

\begin{table}[t!]
\centering
\caption{Ablation of point-refinement steps using Qwen3-VL-8B on the MeViS dataset. We report $\mathcal{J}\&\mathcal{F}$.}
\label{tab:rebuttal_qwen3_refine}
\vspace{2pt}
\footnotesize
\setlength{\tabcolsep}{4.0pt}
\renewcommand{\arraystretch}{1.04}
\begin{tabular}{lcccc}
\toprule
\textbf{Qwen3-VL-8B} & \multicolumn{4}{c}{\textbf{Training steps}} \\
\cmidrule(lr){2-5}
\textbf{Test steps} & $R=1$ & $R=2$ & $R=3$ & $R=4$ \\
\midrule
$R=1$ & 55.1 & 56.8 & 57.7 & 58.0 \\
$R=2$ & 59.1 & 60.4 & 62.5 & \textbf{62.8} \\
\bottomrule
\end{tabular}
\end{table}

\begin{table}[t!]
\centering
\captionsetup{aboveskip=2pt, belowskip=2pt}
\caption{Ablation study on the number of points $K$ in the SPGR module.}
\label{tab:abl_point_num}
\vspace{2pt}
\footnotesize
\setlength{\tabcolsep}{5.0pt}
\renewcommand{\arraystretch}{1.04}
\begin{tabular*}{0.48\textwidth}{@{\extracolsep{\fill}}lccc@{}}
\toprule
\textbf{$K$} & $\mathcal{J}$ & $\mathcal{F}$ & $\mathcal{J}\&\mathcal{F}$ \\
\midrule
1 & 53.6 & 58.4 & 56.0 \\
3 & 56.9 & 61.6 & 59.2 \\
5 & 57.5 & 62.4 & 60.0 \\
7 & \textbf{57.7} & 62.5 & \textbf{60.1} \\
9 & \textbf{57.7} & \textbf{62.6} & \textbf{60.1} \\
\bottomrule
\end{tabular*}
\vspace{-3mm}
\end{table}

\noindent\textbf{Multi-instance segmentation.}
Our method is able to handle cases where a single query refers to multiple object instances. Specifically, for a multi-instance query, we regard all instances as a target set and define the ground truth mask as the union of their individual masks. Our model still predicts one bounding box and one point set  for the target set. For box supervision, we use the enclosing box of all target instances as the ground truth box. For point supervision, each predicted point is regarded as valid as long as it falls inside any of the referred instance masks. SAM2 then treats the box and all point prompts jointly as spatial guidance and predicts the masks for target instances. Thus, no extra modification is required for multi-instance expressions. 
We report results on the ReVOS dataset, which contains 4,719 single-instance samples and 880 multi-instance samples. As shown in Table~\ref{tab:revos_detailed}, our method achieves an overall J\&F of 66.5\% on single-instance samples and 64.6\% on multi-instance samples. They are not much different. As shown in Fig.~\ref{multiinstance}, our method can effectively handle multi-instance expressions.

\noindent\textbf{The point refinement on Qwen3-VL-8B.}
We replace the default Qwen2.5-VL backbone of our model with stronger Qwen3-VL-8B. As shown in Table~\ref{tab:rebuttal_qwen3_refine}, point refinement  consistently improves performance across different Qwen-VL backbones. In particular, increasing the number of refinement steps ($R$ in the table) during either training or inference leads to performance gains for Qwen3-VL-8B.

\noindent\textbf{The number of predicted points.} 
We further verify the effect of different number of points in the predicted point set $K$ in the Spatial-aware Prompt Generation and Refinement (SPGR) module. As shown in Tab.~\ref{tab:abl_point_num}, 
increasing $K$ from 1 to 7 consistently improves the performance, while the performance becomes stable when $K\geq7$.
Therefore, we set $K=7$ as default.

\begin{figure*}[!t]
\centering
\captionsetup{aboveskip=10pt, belowskip=5pt}
\includegraphics[width=\linewidth]
{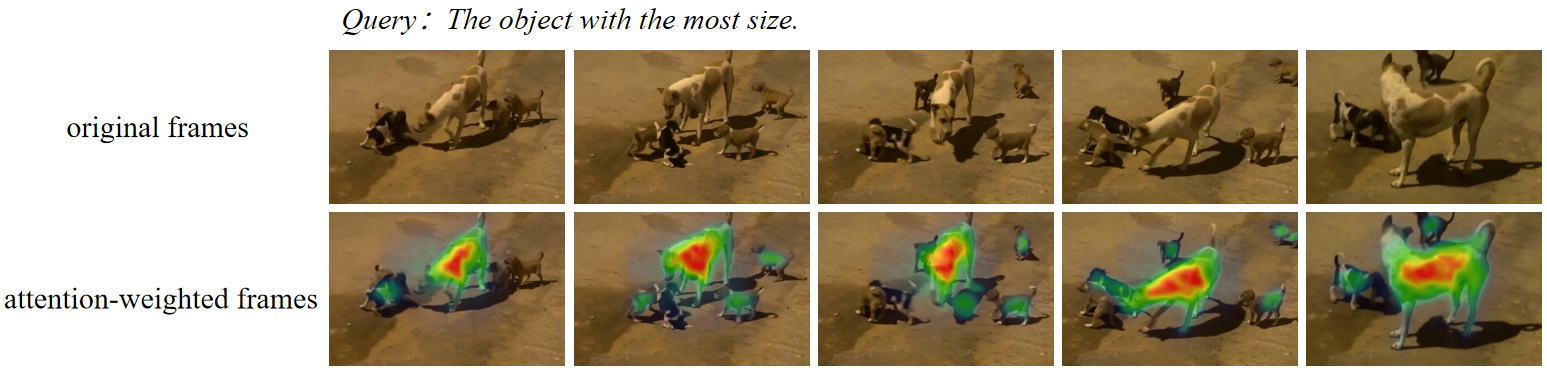}
   \caption{Visualization of attention-weighted results. }
\label{attention}
\vspace{-5mm}
\end{figure*}

\begin{table}[t]
\centering
\caption{Comparison of different point set distribution methods on ReVOS dataset.}
\label{tab:gmm_ablation}
\footnotesize
\setlength{\tabcolsep}{4pt}
\begin{tabular}{@{}lccc@{}}
\toprule
Method & Referring & Reasoning & Overall \\
\midrule
Ours w/ GMM (7 points)& 68.5 & 61.5 & 65.0 \\
Ours w/ GMM (21 points)& \textbf{69.2} & \textbf{63.4} & \textbf{66.3} \\
Ours        & 69.1 & 63.3 & 66.2 \\
\bottomrule
\end{tabular}
\end{table}

\subsection{Query Diversification and Alignment}

\noindent\textbf{Analysis of the Gaussian distribution assumption.}
In our design, we utilize a single Gaussian distribution to represent the point distribution. Previous works such as KLD~\cite{kld} and ProbIOU~\cite{probiou} demonstrate that a single Gaussian distribution can effectively characterize common objects. Although the reasoning targets involve complex, non-rigid, or articulable objects, as reported in~\cite{qwen2.5vl}, the foundational vision-language capabilities of Qwen2.5-VL-7B are primarily learned from large-scale image-text pairs through image-level semantic alignment, leading the model to focus on semantically salient object regions. When we ask the Qwen model to ground the object with a few points, the points are thus most likely from the salient region of the object. 
We show an example in Figure~\ref{attention}, for a dog with large limb motion, the MLLM's attended regions are mainly concentrated around the main body part, while the limbs receive much weaker attention. This attention pattern suggests that a single Gaussian distribution is sufficient to capture the main attended region of the target object.

Furthermore, we also consider using Gaussian Mixture Models (GMMs) to better model the potential target, which meanwhile introduce additional hyperparameters, including the number of mixture components, the number of EM iterations and the number of Monte-Carlo samples for divergence estimation. 
Theoretically, the appropriate number of Gaussians may even vary across objects, making it difficult to determine a universally optimal configuration. 
To examine whether a more expressive distribution improves performance, we replace our default single-Gaussian model with a three-component GMM. Each component is represented by 7 points, making a point set of 21 points. 
Specifically, given the point set, we first use k-means clustering to obtain three spatial groups and use these groups to initialize the mixture components. We then run five EM iterations to estimate the mixture weights, component means, and full covariance matrices. To compute the GMM consistency loss, we follow~\cite{kld} to approximate the symmetric KL divergence with Monte-Carlo sampling. 
As shown in Table~\ref{tab:gmm_ablation}, the GMM achieves $66.3$\% J\&F, which is slightly higher than the $66.2$\% J\&F obtained by our default setup. Nevertheless, we notice that our single Gaussian uses 7 points in total, while the GMM uses 21 points. If we stick to only 7 points for the GMM, the overall performance drops significantly to $65.0$\% J\&F. 
Therefore, although a single Gaussian may be less precise in a few cases, it provides a more favorable trade-off between simplicity, efficiency, and accuracy. 

Next, we argue that different reasoning queries are designed to highlight different regions of the same target object. 
Strict point-wise matching methods, such as the Chamfer distance and Earth Mover's Distance, impose strong correspondence constraint between the two point sets, which can be overly restrictive for our setting. In fact, our ablation in Table~8 in the paper confirms this, \ie, replacing $\mathcal{L}_p$ with nearest-point matching ($\mathcal{L}_{p\text{-strict}}$) decreases performance from $60.1$\% to $57.6$\% J\&F.

Finally, we optimize both the means and covariance matrices of the Gaussian distributions via the symmetric KL divergence point consistency loss, as detailed in Sec~3.2 (Eq. 5) in the paper. Therefore, it is unlikely that the actual activated regions from the two distributions are completely disjointed when we jointly align their means and variances.




\begin{table}[t!]
\centering
\caption{Method comparison with Qwen2.5-VL-7B backbone and SAM2 decoder on ReVOS dataset.}
\label{tab:comparison_revos}
\footnotesize
\setlength{\tabcolsep}{4pt}
\begin{tabular}{@{}llcccc@{}}
\toprule
Method & Backbone &decoder & Referring & Reasoning & Overall \\
\midrule
LISA~\cite{lisa} & Qwen2.5-VL-7B &SAM2& 50.2 & 40.8 & 45.5 \\
VISA~\cite{visa} & Qwen2.5-VL-7B &SAM2& 55.6 & 49.7 & 52.7 \\
Ours & Qwen2.5-VL-7B &SAM2& 69.1 & 63.3 & 66.2 \\
\bottomrule
\end{tabular}
\end{table}

\subsection{Additional study regarding baselines}
Tables~1 and~2 in the paper already include some state-of-the-art methods that utilize the same Qwen2.5-VL-7B backbone and SAM2 decoder as our model, \eg, UniPixel~\cite{unipixel} (NeurIPS 2025), SDAM~\cite{sdam} (AAAI 2026), VideoSeg-R1 ~\cite{videosegr1} (AAAI 2026), and DecAF ~\cite{decaf} (ICLR 2026). Our method consistently outperforms them.
Moreover, as shown in Table~\ref{tab:comparison_revos}, we replace the original MLLM backbones of LISA~\cite{lisa} and VISA~\cite{visa} with Qwen2.5-VL-7B and replace their original decoders with SAM2 to enable a fair comparison to ours. We can see our method still outperforms VISA by 13.5\% and LISA by 20.7\% in J\&F on ReVOS dataset.

\begin{figure*}[t!]
\centering
\captionsetup{aboveskip=10pt, belowskip=5pt}
\includegraphics[width=\linewidth]
{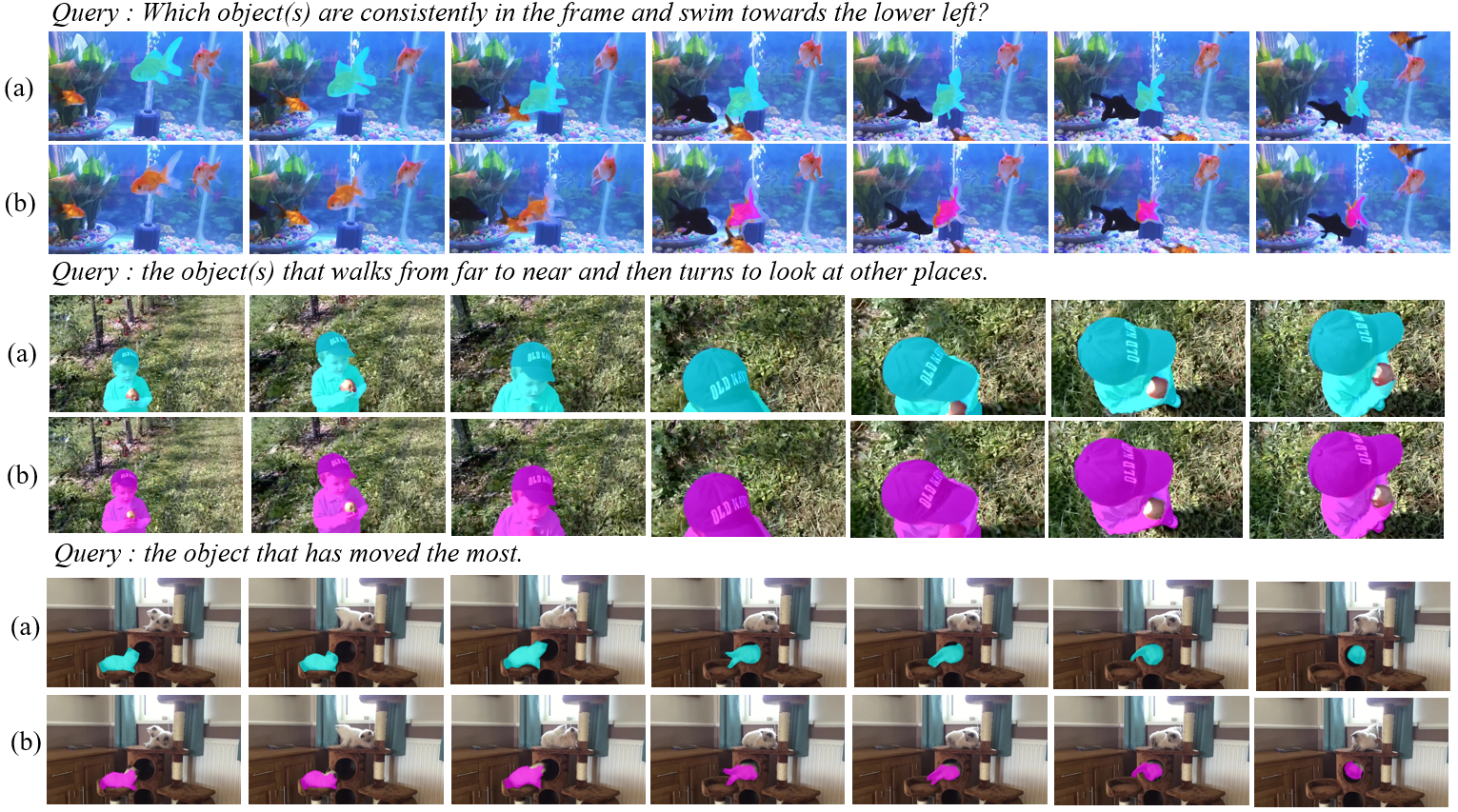}
   \caption{More Visualization results on ReVOS. (a) Our method, (b) UniPixel~\cite{unipixel}. }
\label{more vis}
\vspace{-5mm}
\end{figure*}

\begin{figure*}[t!]
\centering
\captionsetup{aboveskip=10pt, belowskip=5pt}
\includegraphics[width=\linewidth]
{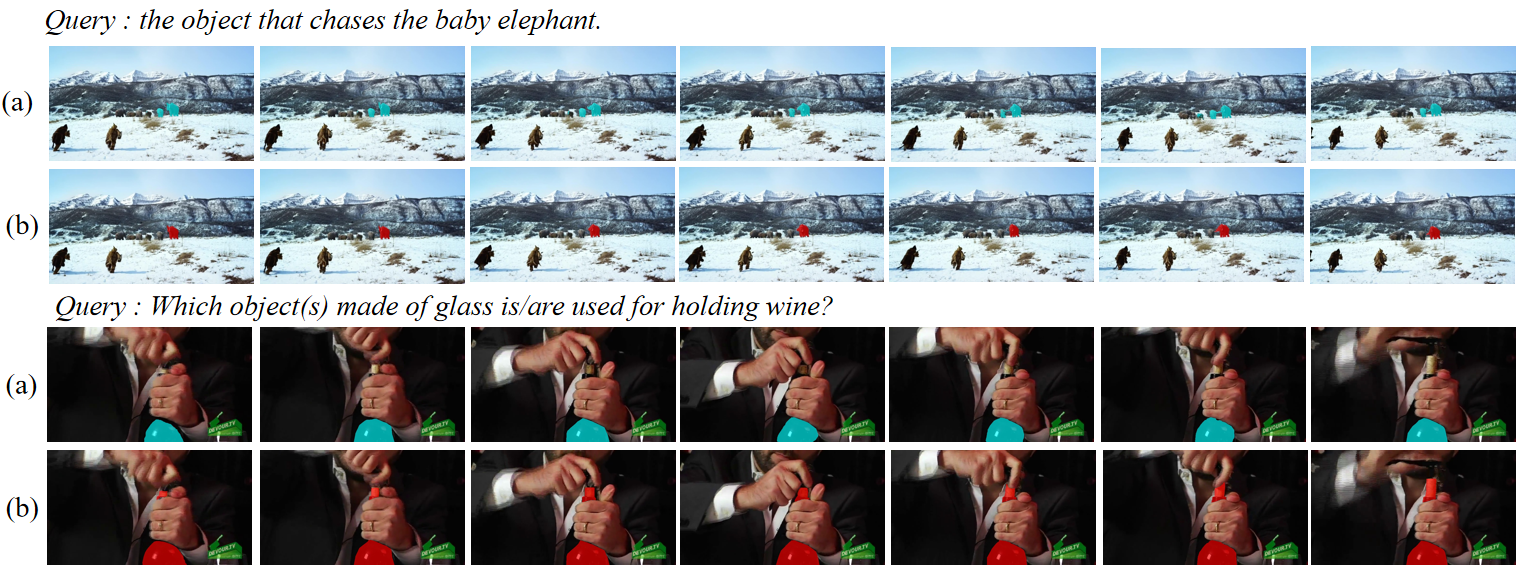}
   \caption{Failure cases on ReVOS. (a) Our method, (b) ground truth. }
\label{failure}
\vspace{-5mm}
\end{figure*}
\section{More Visualization Results} We present additional qualitative results of our method and a comparison method, \ie, UniPixel~\cite{unipixel} across video frames in Fig.~\ref{more vis}.

\section{Failure Cases} We present some failure cases in Fig.~\ref{failure} and summarize the key limitations. First, when numerous instances from the same or visually similar categories appear simultaneously, our method may struggle to distinguish the target. Second, our method may fail when the target object is heavily occluded.


\newpage
\section*{NeurIPS Paper Checklist}

\begin{enumerate}

\item {\bf Claims}
    \item[] Question: Do the main claims made in the abstract and introduction accurately reflect the paper's contributions and scope?
    \item[] Answer: \answerYes{} 
    \item[] Justification: The main claims made in the abstract and introduction accurately reflect the paper's contributions and scope.

\item {\bf Limitations}
    \item[] Question: Does the paper discuss the limitations of the work performed by the authors?
    \item[] Answer: \answerYes{} 
    \item[] Justification: We discuss the failure cases in the appendix.

\item {\bf Theory assumptions and proofs}
    \item[] Question: For each theoretical result, does the paper provide the full set of assumptions and a complete (and correct) proof?
    \item[] Answer: \answerNA{} 
    \item[] Justification: The paper does not include theoretical results. 

    \item {\bf Experimental result reproducibility}
    \item[] Question: Does the paper fully disclose all the information needed to reproduce the main experimental results of the paper to the extent that it affects the main claims and/or conclusions of the paper (regardless of whether the code and data are provided or not)?
    \item[] Answer: \answerYes{} 
    \item[] Justification: The paper fully discloses all the information needed to reproduce the main experimental results.

\item {\bf Open access to data and code}
    \item[] Question: Does the paper provide open access to the data and code, with sufficient instructions to faithfully reproduce the main experimental results, as described in supplemental material?
    \item[] Answer: \answerYes{} 
    \item[] Justification: The full code has been released after being accepted.

\item {\bf Experimental setting/details}
    \item[] Question: Does the paper specify all the training and test details (e.g., data splits, hyperparameters, how they were chosen, type of optimizer) necessary to understand the results?
    \item[] Answer: \answerYes{} 
    \item[] Justification: We specify all the training and test details in the implementation details section.

\item {\bf Experiment statistical significance}
    \item[] Question: Does the paper report error bars suitably and correctly defined or other appropriate information about the statistical significance of the experiments?
    \item[] Answer: \answerNo{} 
    \item[] Justification: We do not report the information about the statistical significance of the experiments.

\item {\bf Experiments compute resources}
    \item[] Question: For each experiment, does the paper provide sufficient information on the computer resources (type of compute workers, memory, time of execution) needed to reproduce the experiments?
    \item[] Answer: \answerYes{} 
    \item[] Justification: We report the computer resources in Section Experiments.
    
\item {\bf Code of ethics}
    \item[] Question: Does the research conducted in the paper conform, in every respect, with the NeurIPS Code of Ethics \url{https://neurips.cc/public/EthicsGuidelines}?
    \item[] Answer: \answerYes{} 
    \item[] Justification: Our research conforms with the NeurIPS Code of Ethics.

\item {\bf Broader impacts}
    \item[] Question: Does the paper discuss both potential positive societal impacts and negative societal impacts of the work performed?
    \item[] Answer: \answerNo{} 
    \item[] Justification: We do not discuss the potential positive societal impacts and negative societal impacts of the work.
    
\item {\bf Safeguards}
    \item[] Question: Does the paper describe safeguards that have been put in place for responsible release of data or models that have a high risk for misuse (e.g., pre-trained language models, image generators, or scraped datasets)?
    \item[] Answer: \answerNA{} 
    \item[] Justification: The paper poses no such risks.

\item {\bf Licenses for existing assets}
    \item[] Question: Are the creators or original owners of assets (e.g., code, data, models), used in the paper, properly credited and are the license and terms of use explicitly mentioned and properly respected?
    \item[] Answer: \answerYes{} 
    \item[] Justification: We have properly credited all original owners of assets.

\item {\bf New assets}
    \item[] Question: Are new assets introduced in the paper well documented and is the documentation provided alongside the assets?
    \item[] Answer: \answerNA{} 
    \item[] Justification: The paper does not release new assets.

\item {\bf Crowdsourcing and research with human subjects}
    \item[] Question: For crowdsourcing experiments and research with human subjects, does the paper include the full text of instructions given to participants and screenshots, if applicable, as well as details about compensation (if any)? 
    \item[] Answer: \answerNA{} 
    \item[] Justification: The paper does not involve crowdsourcing nor research with human subjects.

\item {\bf Institutional review board (IRB) approvals or equivalent for research with human subjects}
    \item[] Question: Does the paper describe potential risks incurred by study participants, whether such risks were disclosed to the subjects, and whether Institutional Review Board (IRB) approvals (or an equivalent approval/review based on the requirements of your country or institution) were obtained?
    \item[] Answer: \answerNA{} 
    \item[] Justification: The paper does not involve crowdsourcing nor research with human subjects.

\item {\bf Declaration of LLM usage}
    \item[] Question: Does the paper describe the usage of LLMs if it is an important, original, or non-standard component of the core methods in this research? Note that if the LLM is used only for writing, editing, or formatting purposes and does \emph{not} impact the core methodology, scientific rigor, or originality of the research, declaration is not required.
    \item[] Answer: \answerYes{} 
    \item[] Justification: The paper describes the usage of LLMs in detail.

\end{enumerate}

\end{document}